\documentclass[sw]{iosart2x}

\makeatletter
\let\numberlines@hook\relax
\makeatother

\usepackage[russian,english]{babel}
\usepackage{times}
\usepackage{latexsym}
\usepackage{booktabs}
\usepackage{multirow}
\usepackage{enumitem}
\usepackage{float}
\usepackage{afterpage}
\usepackage{subfig}
\usepackage{graphicx}
\usepackage[table,xcdraw]{xcolor}
\usepackage[normalem]{ulem}
\useunder{\uline}{\ul}{}
\usepackage{times}
\usepackage{url}
\usepackage{subfig}
\usepackage{minibox}
\usepackage{longtable}

\DeclareRobustCommand{\cyrins}[1]{%
  \begingroup\fontfamily{erewhon-TLF}%
  \foreignlanguage{russian}{#1}%
  \endgroup
}

\pubyear{0000}
\volume{0}
\firstpage{1}
\lastpage{1}

\begin{document}
\makeatletter
\let\put@numberlines@box\relax
\makeatother
\begin{frontmatter}

\title{Taxonomy Enrichment with Text and Graph Vector Representations
}
\runtitle{Taxonomy Enrichment with Text and Graph Vector Representations}

\author[**]{\inits{I.}\fnms{Irina} \snm{Nikishina}\ead[label=e1]{irina.nikishina@skoltech.ru}%
\thanks{Equal contribution with Mikhail Tikhomirov. Corresponding author: \printead{e1}.}},
\author[*]{\inits{M.}\fnms{Mikhail} \snm{Tikhomirov}\ead[label=e3]{tikhomirov.mm@gmail.com}},
\author[**]{\inits{L.}\fnms{Varvara} \snm{Logacheva}\ead[label=e2]{v.logacheva@skoltech.ru}},
\author[]{\inits{Y.}\fnms{Yuriy} \snm{Nazarov}\ead[label=e4]{nazarov.yuriy.pavlovich@gmail.com}},
\author[**]{\inits{A.}\fnms{Alexander} \snm{Panchenko}\ead[label=e5]{a.panchenko@skoltech.ru}},
and
\author[*]{\inits{N.}\fnms{Natalia} \snm{Loukachevitch}\ead[label=e6]{louk\_nat@mail.ru}}
\address[**]{\orgname{Skolkovo Institute of Science and Technology},
Moscow, \cny{Russia}\printead[presep={\\}]{e1,e2,e5}}
\address[*]{Research Computing Center, \orgname{Lomonosov Moscow State University},
Moscow, \cny{Russia}\printead[presep={\\}]{e3,e4,e6}}

\begin{abstract}

Knowledge graphs such as DBpedia, Freebase or Wikidata always contain a taxonomic backbone that allows the arrangement and structuring of various concepts in accordance with hypo-hypernym (``class-subclass'') relationship. With the rapid growth of lexical resources for specific domains, the problem of automatic extension of the existing knowledge bases with new words is becoming more and more widespread. In this paper, we address the problem of \textit{taxonomy enrichment} which aims at adding new words to the existing taxonomy.

We present a new method which allows achieving high results on this task with little effort. It uses the resources which exist for the majority of languages, making the method universal. We extend our method by incorporating deep representations of graph structures like node2vec, Poincaré embeddings, GCN  etc. that have recently demonstrated promising results on various NLP tasks. Furthermore, combining these representations with word embeddings allows us to beat the state of the art.

We conduct a comprehensive study of the existing approaches to taxonomy enrichment based on word and graph vector representations and their fusion approaches. We also explore the ways of using deep learning architectures to extend taxonomic backbones of knowledge graphs. We create a number of datasets for taxonomy extension for English and Russian. We achieve state-of-the-art results across different datasets and provide an in-depth error analysis of mistakes.

\end{abstract}


\end{frontmatter}

\section{Introduction}

The central idea of Semantic Web is to make the content of the Internet pages machine-interpretable. For that,  web pages should be linked to \textit{ontologies}~\cite{berners2001semantic,gomez2002ontology} --- databases which contain the information on classes of objects, their properties, and relations between the classes. The relations between objects are particularly important in an ontology, because they form its structure. They can be of different types corresponding to different types of relationships between real-world objects. One of the most important relationships is the class-subclass relation. It allows organising entities into a \textit{taxonomy} --- a tree structure where entities are represented as nodes and the edges between them denote \textit{subclass-of} or \textit{instance-of} relationship. The class-subclass relations and taxonomies built from them are crucial for understanding the place and the purpose of an object or a concept in the world. This relation and the hierarchical structure created by it are also a basis of many knowledge bases and \textit{knowledge graphs} --- a particular type of knowledge bases where objects are organised in a graph structure. There, the nodes of a graph are objects, and the edges of a graph are relations between the objects. 

To support the use of specific knowledge base structures, such as thesauri or taxonomies, there exist specifications and standards classification schemes. 
For instance, Simple Knowledge Organization Systems (SKOS)\footnote{\url{https://www.w3.org/2004/02/skos/intro}} define several types of lexical-semantic relations in the terms of semantic web, e.g. ``has broader/narrower'', ``has exact match'', and ``is in mapping relation with''. SKOS are used to create structures like thesauri.

However, SKOS does not fully comply with our taxonomy. The class-subclass relation which is the base of taxonomies is a relation between objects X and Y such that the sentence ``X is a kind of Y'' is acceptable for native speakers. On the other hand, the closest SKOS analogue of taxonomic class-subclass relation is the ``broader term relation''. It is different from the class-subclass relation, because it is less specific. For example, it can include the part-whole relations.


The usefulness of an ontology or a knowledge base depends largely on its completeness and its ability to fully reflect the real world. However, since the world is changing, the ontologies need to be constantly updated to stay relevant. There currently exist comprehensive knowledge bases such as Freebase, DBPedia or Wikidata as well as ontologies for specific domains. Many areas of knowledge require their own knowledge bases, and all of them need to be maintained and extended. This is an expensive and time-consuming process which can only be conducted by an expert who is proficient in the discipline and understands the structure of a knowledge base. Thus, in order to speed up and simplify this task, it becomes more and more important to develop systems that could automatically enrich the existing knowledge bases with new words or at least facilitate the manual extension process. The task of automatically or semi-automatically adding new entities to hierarchical structures is referred to as \textit{taxonomy enrichment}.

In this work, we aim at reviewing the existing taxonomy enrichment models and propose new methods which address their drawback. We also aim at evaluating the scalability of different methods to new languages and datasets.

The state-of-the-art taxonomy enrichment methods have two main drawbacks. First of all, they often use unrealistic formulations of the task. For example, SemEval-2016 task 14~\cite{jurgens-pilehvar-2016-semeval} which was the first effort to evaluate this task in a controlled environment, provided definitions of the query words (words to be added to a taxonomy). This is very informative resource, so the majority of the presented methods heavily depended on those definitions~\cite{tanev-rotondi-2016-deftor,espinosa-anke-etal-2016-taln}. However, in the real-world scenarios, such information is usually unavailable, which makes the developed methods inapplicable. We tackle this problem by testing our new methods and the state-of-the-art methods in a realistic setting.

Another gap in the existing research is that the majority of methods use the information from only one source. Namely, some researchers use the information from distributional word embeddings, whereas others consider graph-based models which represent a word based on its position in a taxonomy. Our intuition is that the information from these two sources is complementary, so combining them can improve the performance of taxonomy enrichment models. Therefore, we propose a number of ways to incorporate various sources of information.

First, we propose the new \textbf{DWRank} method which uses only distributional information from pre-trained word embeddings and is similar to other existing methods. We then enable this method to incorporate the different sources of graph information. We compare the various ways of getting the information from a knowledge graph. Finally, another modification of our method successfully combines the information from different sources, beating the current state of the art.

To place our models in the context of the research on taxonomy enrichment, we compare them with a number of state-of-the-art models. To the best of our knowledge, this is the first large-scale evaluation of taxonomy enrichment methods. We are also the first to evaluate the methods on datasets of different sizes and in different languages.

This work is an extended version of the work described in \cite{nikishina-etal-2020-studying,nikishina-etal-2021-exploring, tikhomirov2021meta}. The novelty of this particular article as compared to the previous publications is as follows:

\begin{enumerate}
\item We present a new taxonomy enrichment method \textbf{DWRank} which combines distributional information and the information extracted from Wiktionary.

\item We present an extension of DWRank called \textbf{DWRank-Graph} which uses various graph-based representations via a common interface.

\item We present \textbf{DWRank-Meta} --- an extension of DWRank which combines the information from different sources and beats the state-of-the-art models.

\item We present \textbf{WBSR} --- a method for taxonomy extension which leverages the information from the Web. This approach is the current state of the art in the task.

\item We conduct a large-scale computational study of various approaches to taxonomy enrichment, which features multiple methods (including ours as well as state-of-the-art approaches), multiple datasets and languages.

\item We present datasets for studying the 
evolution of wordnets for English and Russian, extending the monolingual setup of the RUSSE'2020 shared task \cite{nikishina2020taxonomy} with a larger Russian dataset and similar English versions.

\item We explore the benefits of meta-embeddings (combinations of embeddings) and graph embeddings for the task of taxonomy enrichment.

\item We provide an in-depth error analysis of different types of mistakes of the state-of-the art models.

\item We provide mappings to WordNet Linked Open Data (LOD) Inter-Lingual Index (ILI) from Russian to English synsets. A dataset of multilingual hypernyms  opens possibilities for various use-cases for cross-lingual operations on taxonomies. \end{enumerate}

We release all our code and datasets to enable further research in this direction.\footnote{\url{https://github.com/skoltech-nlp/diachronic-wordnets}} 

The rest of the paper is organised as follows. In Section~\ref{sec:formulation} we formulate the task and provide the relevant definitions. Section~\ref{sec:related} is devoted to the existing work on taxonomy enrichment. Then, in Section~\ref{sec:dataset} we present our datasets and describe their creation process and features. Sections~\ref{sec:baselines},~\ref{sec:graph}, and~\ref{sec:dwrank_meta} contain the descriptions of our methods: in Section~\ref{sec:baselines} we introduce our core method DWRank and in Sections~\ref{sec:graph} and~\ref{sec:dwrank_meta} describe its extensions DWRank-Graph and DWRank-Meta. We report the performance of our models and compare them with the other approaches in Section~\ref{sec:evaluation}. Finally, in Section~\ref{sec:error_analysis} we analyse the potential limitations of our methods. 

\section{Task Formulation and Definitions}
\label{sec:formulation}

In this subsection we formulate the task and
explain the main concepts and task-related terms.

\subsection{Task Formulation}

\begin{figure}[ht]
    \centering
    \caption{Example of adding a new word ``Papuan'' to the taxonomy}
    \label{example:wn}
    \includegraphics[width=0.35\textwidth]{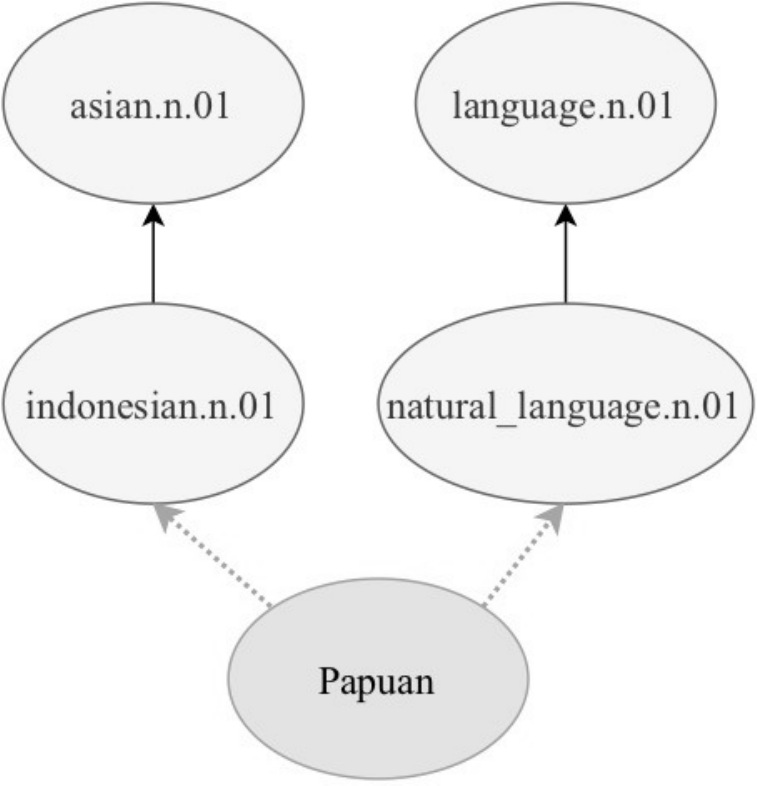}
\end{figure}

Let us consider an example of taxonomy. 
Figure \ref{example:wn} demonstrates a subgraph for the word ``Papuan'' retrieved from WordNet. There, each concept has one or more \textit{parents} (concepts which it is derived from). The parents in their turn have their own parents, and one can trace the word affiliation until the most abstract \textit{root} concepts of the taxonomy. Here, the word ``Papuan'' is attached to the synsets ``indonesian'' and ``natural\_language'' as it can mean both an ethnicity and a language. Note that words can have multiple parents for different reasons. Two or more parents can point to the fact that a word has multiple meanings (as in the case with ``Papuan''). Alternatively, a word meaning can be a combination of meanings of two higher-level concepts.

Therefore, in order to add a word to a taxonomy, we need to find its hypernym among the entities (synsets in case of wordnets) of this taxonomy. Here, we refer to a word absent from the taxonomy (a word which we would like to add) as a \textit{query word}. Our task is to attach query words to an existing taxonomic tree.

The task of finding a single suitable hypernym synset is difficult for a machine, because the number of nodes in the existing taxonomy can be very large (e.g. in WordNet the number of noun synsets is around 29,000). Thus, a model trained to solve this task will inevitably return many false answers if asked to provide only one synset candidate. Moreover, as we can see from Figure~\ref{example:wn}, a word may have multiple hypernyms.

Thus, in the majority of works on taxonomy enrichment the requirement of providing a single correct answer is relaxed. Instead of that, a common approach is to provide $k$ (typically 10 to 20) most suitable candidates. This list is more likely to contain correct synsets. This setting is also consistent with the manual computer-assisted annotation. Presenting an annotator with a small list of candidates will facilitate the taxonomy extension process: annotators will only have to look through a short list of high-probability hypernym candidates instead of searching hypernyms in a list of all synsets of the taxonomy. Thus, we formulate the task of \textit{taxonomy extension} as a soft ranking problem, where the synsets are ranked according to their suitability for a given word.

This is an established approach to this task, the same formulation was also used in research works on taxonomy enrichment and in taxonomy enrichment shared tasks~\cite{jurgens-pilehvar-2016-semeval,nikishina2020taxonomy}.

\subsection{Definitions}
In this section, we provide several key definitions crucial for understanding of our work.

\paragraph{\textbf{Knowledge base}} is a data structure used to store knowledge which comprise of concepts and relations (known as \textsf{TBox}-es) and individuals, concept, and relation instantiations (also known as \textsf{ABox}-es)~\cite{TAboxes}.

\paragraph{\textbf{Knowledge graph}} is a knowledge base that uses a graph-structured data model to represent data. Formally, a knowledge graph $G$ is a set of triplets $\{h, r, t\} \subseteq E \times R \times E$ where $E$ is the entity set and $R$ is the relation set. Normally, knowledge graphs comprise multiple types of relations $R$.

\paragraph{\textbf{Synset}} is a set $S \subseteq \{s_1, ..., s_n\}$ of words or phrases corresponding to a concept or a knowledge base entity in the set $E$. Synset is the major element which denotes a node in a knowledge graph $G$, therefore, in our case $S$ equals $E$.

\paragraph{\textbf{Hypernymy relation}} is a relation $r \in R$. According to \cite{miller1998wordnet}, hypernymy relation exists between objects X and Y if native speakers  accept sentences constructed using  such patterns as ``An X is a (kind of) Y''. Hypernymy is transitive and asymmetrical. Such relations are central organizing relations for  nouns and verbs in WordNet \cite{miller1998wordnet}. In fact, hypernymy relations comprise class (or subsumption) relations and instantiation relations considered in ontology studies and description logics \cite{mcguinness1995explaining}.  

\paragraph{\textbf{Taxonomy}} is a special case of a knowledge graph $G$.
It is a tree-based structure where nodes from the set $E$ are connected with the hypernymy relation $r$. Thus, the set of relations $R = \{r\}$. Elements included in the set $E$ are words or concepts. In a taxonomy $G$ they are arranged in a hierarchical structure from the most abstract concept (\textit{root}) to the most specific concepts (\textit{leaves}).


\paragraph{\textbf{Query words} (new words)} --- words to be added to the taxonomy $T$, usually manually collected by experts. In this paper, we use the following notation for this set $Q \subseteq \{q_1, ..., q_n\}$. Note that words may be ambiguous: one $q_i$ can correspond to more than one synset $s_i$.

\paragraph{\textbf{Knowledge graph completion}} is a task of completing a Knowledge Graph $G$ by finding a set of missing triples $\{ h, r, t | h~\in~E, r~\in~R, t~\in~E, ~h,~r,~t~~\notin~G\}$.

\paragraph{\textbf{Taxonomy enrichment}} can be considered as a special case of the knowledge base completion task. This task aimed at associating each new word $q \in Q$, which is not yet included in the taxonomy $T$, with the appropriate hypernyms from it. 

\section{Related Work}
\label{sec:related}

There exist numerous approaches to knowledge base completion, which make use of various neural network architectures described in review papers \cite{kbcompletion,paulheim2017knowledge}. Most of them apply low-dimensional graph embeddings \cite{transe,proje}, deep learning architectures like autoencoders \cite{autoencoder} or graph convolutional networks \cite{li2019ontology,dettmers2018convolutional,shang2019end}. Another group of approaches makes use of tensor decomposition approaches: Tucker decomposition \cite{balazevic-etal-2019-tucker} and Canonical Polyadic decomposition (CANDECOMP/PARAFAC) \cite{pmlr-v80-lacroix18a,Lacroix20}. 
Another task, which is closely related to the taxonomy creation is the Knowledge base construction. There exists multiple approaches solving the problem: Text2Onto \cite{text2onto} a pillar language-independent approach which applies user interaction or fully automated methods \cite{DESSI2021253,klink2} which apply text-mining tools and external sources, which are mostly applied for the scholarly domain. However, knowledge base completion task assumes a generic graph while in the taxonomy enrichment task deals with tree structures and specific methods of tree processing are commonly used in this field. In this article we focus on this specific task and narrow down our scope to enrichment and population of taxonomic structures. It should be noted however that the majority of the ontologies and knowledge bases possess some kind of taxonomic backbone and therefore the task of construction and maintaining such a semantic structure is fundamental.

The existing studies on the taxonomies can be divided into three groups. The first one addresses the Hypernym Discovery problem \cite{camacho-collados-etal-2018-semeval}: given a word and a text corpus, the task is to identify hypernyms in the text. However, in this task the participants are not given any predefined taxonomy to rely on. The second group of works deals with the Taxonomy Induction problem \cite{bordea-etal-2015-semeval,bordea-etal-2016-semeval,velardi-etal-2013-ontolearn}, in other words, creation of a taxonomy from scratch. 
Finally, the third direction of research is the Taxonomy Enrichment task: the participants extend a given taxonomy with new words. In this article, we focus on the taxonomy enrichment task and explore various approaches based on text and graph embeddings and their combinations to the solution of this problem. The following sections provide overview of the various strains of research related to the given problem. 

\subsection{Prior Art on Taxonomy Enrichment}

Until recently, the only dataset for the taxonomy enrichment task was created under the scope of SemEval-2016. It contained definitions for the new words, so the majority of models solving this task used the definitions. For instance, \textit{Deftor} team~\cite{tanev-rotondi-2016-deftor} computed definition vector for the input word, comparing it with the vector of the candidate definitions from WordNet using cosine similarity. Another example is \textit{TALN} team~\cite{espinosa-anke-etal-2016-taln} which also makes use of the definition by extracting noun and verb phrases for candidates generation.

This scenario may be unrealistic for manual annotation because annotators are writing a definition for a new word and adding new words to the taxonomy simultaneously. 
Having a list of candidates would not only speed up the annotation process but also identify the range of possible senses. Moreover, it is possible that not yet included words may have no definition in any other sources: they could be very rare (``apparatchik'', ``falanga''), relatively new (``selfie'', ``hashtag'') or come from a 
narrow domain (``vermiculite'').

Thus, following RUSSE-2020 shared task~\cite{nikishina2020taxonomy}, we stick to a more realistic scenario when we have no definitions of new words, but only examples of their usage. For this shared task we provide a baseline as well as training and evaluation datasets based on RuWordNet~\cite{loukachevitch2016creating} which will be discussed in the next section. The task exploited words which were recently added to the latest release of RuWordNet and for which the hypernym synsets for the words were already identified by qualified annotators. The participants of the competition were asked to find synsets which could be used as hypernyms.  

The participated systems mainly relied on vector representations of words and the intuition that words used in similar contexts have close meanings. They cast the task as a classification problem where words need to be assigned one or more hypernyms \cite{kunilovskaya2020russe} or ranked all hypernyms by suitability for a particular word \cite{dale2020russe}. They also used a range of additional resources, such as Wiktionary, dictionaries, additional corpora  \cite{arefyev2020russe}.
Interestingly, only one of the well-performing models \cite{tikhomirov2020russe} used context-informed embeddings (BERT) or external tools such as online Machine Translation (MT) and search engines (the best-performing model denoted as \textit{Yuriy} in the workshop description paper). 

In this paper, we would like to work out methods which depend on graph based structures and combine them with the  approaches applying word embeddings. At the same time, we want our methods to benefit from the existing data (e.g. corpora, pre-trained embeddings, Wiktionary).

\subsection{Word Vector Representations for Taxonomies} 

Approaches using word vector representations are the most popular choice for all tasks related to taxonomies. When solving the \textit{Hypernym Discovery} problem in SemEval-2018 Task~9 \cite{camacho-collados-etal-2018-semeval} most of participants use word embeddings. For instance, Bernier-Colborne and Barriere~\cite{bernier-colborne-barriere-2018-crim} predict the likelihood of the relationship between an input word and a candidate using word2vec embeddings. Berend et al.~\cite{berend-etal-2018-300} use Word2vec vectors as features of a logistic regression classifier. Maldonado and Klubicka~\cite{maldonado-klubicka-2018-adapt} simply consider top-10 closest associates from the Skip-gram word2vec model as hypernym candidates.
Pre-trained GloVe embeddings~\cite{pennington-etal-2014-glove} are also used to initialize embeddings for an LSTM-based Hypernymy Detection model \cite{shwartz-etal-2016-improving}.

Participants also solve the SemEval-2016 Task~13 on taxonomy induction with word embeddings \cite{pocostales-2016-nuig}: they compute the vector offset as the average offset of all the pairs generated and exploit it to predict hypernyms for the new data. Afterwards, in \cite{aly-etal-2019-every} the authors apply word2vec embeddings similarity to improve the approaches of the SemEval-2016 Task 13 participants.

The vast majority of participants of SemEval-2016 task~14 \cite{jurgens-pilehvar-2016-semeval} and RUSSE'2020 \cite{nikishina2020taxonomy} also apply word embeddings to find the correct hypernyms in the existing taxonomy. For instance, the participants compute a definition vector for the input word by comparing it with the definition vectors of the candidates from the wordnet using cosine similarity \cite{tanev-rotondi-2016-deftor}. Another option is to train word2vec embeddings from scratch and cast the task as a classification problem \cite{kunilovskaya2020russe}. Some participants compare the approach based on XLM-R model \cite{conneau-etal-2020-unsupervised} with the word2vec ``hypernyms of co-hyponyms'' method \cite{arefyev2020russe}. It considers nearest neighbours as co-hyponyms and takes their hypernyms as candidate synsets.

Summing up, the usage of distributed word vector representations is a simple yet efficient approach to the taxonomy-related tasks and 
should be considered a strong baseline \cite{camacho-collados-etal-2018-semeval,nikishina2020taxonomy}.

\subsection{Meta-Embedding Approaches to Word Representation}


Vector representations can be learned on various datasets and using various models. It has been shown that combining word embeddings is beneficial for NLP tasks, e.g. dependency parsing \cite{bansal2014tailoring}, and in medical domain \cite{chowdhury2019mixed}.

Coates et al.~\cite{coates2018} show that simple vector combining approaches, such as concatenation or averaging, can significantly improve the overall performance for several tasks. For instance, singular value decomposition (SVD) demonstrates good results with the ability to control the final dimension of vectors~\cite{yin2016learning}. Autoencoders~\cite{bollegala2018learning} further promote the idea of creating meta-embeddings. The authors propose several algorithms 
for combining various word vectors into one vector by encoding initial vectors into some meta-embedding space and then decoding backwards.  

As for the CAEME approach, all word vectors are encoded into meta-vectors and then concatenated. 
Then, the decoding step uses a concatenated representation to predict the original vector representations. The AAEME approach is similar to CAEME, except that each vector is mapped to a fixed-size vector and all encoded representations are averaged, but not concatenated. An obvious advantage of this approach is the ability to control the dimension of meta-embeddings. 

For any AEME approach, different loss functions can be used at the decoding stage: MSE loss, KL-divergence loss, cosine distance loss and also their combinations. In  \cite{neill2018meta} the authors investigated the performance of the autoencoders depending on the loss function. They discover that there is no evident winner across tasks and that  different loss functions are defined be different tasks.

Meta-embeddings are already used in such machine learning tasks as dependency parsing~\cite{bansal2014tailoring}, classification in healthcare~\cite{chowdhury2019mixed}, named-entity recognition~\cite{winata2019learning,neill2018meta}, sentiment analysis~\cite{neill2018meta}, word similarity and analogy tasks~\cite{coates2018,bollegala2018learning,yin2016learning}. To the best of our knowledge, meta-embeddings have not been applied to the taxonomy enrichment task, especially for the fusion of texts and graph embeddings.

\subsection{Graph-based Representations for Taxonomies} 

Taxonomies can be represented as graphs and there exist various approaches to learn graph-based representations. They are thoroughly compared in~\cite{makarov2021fusion} on the link prediction task, which is closely related to Taxonomy Enrichment. The paper also demonstrates that the combination of text and graph embeddings gives a boost on the link prediction task. Most of those methods listed in~\cite{makarov2021fusion} have also been tested on tasks related to the taxonomy enrichment.

For instance, node2vec embeddings \cite{node2vec-kdd2016} are used for taxonomy induction among other network embeddings \cite{liu2018interpretation}.
In \cite{aly-etal-2019-every}, the authors perform the same task. They use hyperbolic Poincaré embeddings to enhance automatically created taxonomies. The SemEval-2016 subtask of reattaching query words to the taxonomy is quite similar to taxonomy enrichment which we perform. 
However, the datasets of the SemEval-2016 Task~13 are restricted to specific domains, which leaves an open question of the efficiency of Poincaré embeddings for the general domain and larger datasets. Moreover, \cite{aly-etal-2019-every} use Hearst Patterns to discover hyponym-hypernym relationships. This technique operates on words, and cannot be transferred to word-synset relations without extra manipulation.

As for the Knowledge Graph Construction task, which is a more general task in relation to the Taxonomy Induction, the vast majority of approaches also use word embeddings as node representations. Several approaches like \cite{luan-etal-2018-multi} and apply ELMO embeddings \cite{peters-etal-2018-deep} to predict entities and their relations for the knowledge graph. Other approaches \cite{han2020wikicssh} utilize a combination of ELMO, Poincaré and node2vec embeddings to enhance knowledge graph build upon Wikipedia. 

Graph convolutional networks (GCNs) \cite{kipf2016semi} as well as graph autoencoders \cite{kipf2016variational} are mostly applied to the link prediction task on large knowledge bases. For example, in \cite{rossi2020knowledge} the authors present an expanded review of the field and compare a wide variety of existing approaches. Graph embeddings are also often used for other taxonomy-related tasks, e.g. entity linking \cite{pujary2020disease}. As for the taxonomy enrichment task, we are only aware of a recent approach TaxoExpan \cite{shen2020taxoexpan} which applies position-enhanced graph neural networks (GCN \cite{kipf2017semi} and GAT \cite{velickovic2018graph}) that we also evaluate on our datasets\footnote{The results achieved on our datasets are significantly lower than the baseline, probably because of the incorrect model launching.}.

Thus, to the best of our knowledge, our work is the first computational study of Taxonomy enrichment task which aggregates and considers different existing and new approaches for taxonomy enrichment. We compare graph- and word-based representations computed from the synsets and hypo-hypernym relations for hypernym prediction demonstrating state-of-the-art results.

\section{Diachronic WordNet Datasets}
\label{sec:dataset}

The important part of our study is the observation that one can learn from the history of the development of lexical resources though time. More specifically, we make use of the various historic snapshots (versions) of WordNet lexical graphs and setup a task of their automatic completion assuming the manual update of the ground truth. This diachronic analysis -- similar to diachronic lexical analysis of word meanings -- is used to build two datasets in our study: one for English, another one for Russian based respectively on Princeton WordNet \cite{miller1995wordnet} and RuWordNet taxonomies. It is important to mention that by using the word ``diachronic'' we do not imply lexical diachrony, e.g., semantic shifts of words over time \cite{schlechtweg-etal-2020-semeval}, but the temporal extension of Wordnet stored in its versions. Each dataset consists of a taxonomy and a set of novel words to be added to this resource. The statistics are provided in Table \ref{tab:dataset}.

\begin{table}[ht]
\centering
\small
\caption{Statistics of two diachronic WordNet datasets used in this study.}
\label{tab:dataset}
\begin{tabular}{lrr}
\toprule
\textbf{Dataset} & \textbf{Nouns} &\textbf{Verbs} 
\\ \midrule
\textit{WordNet1.6 - WordNet3.0}     & 17 043  & 755 \\
\textit{WordNet1.7 - WordNet3.0}     & 6 161  & 362 \\
\textit{WordNet2.0 - WordNet3.0}     & 2 620  & 193 \\
\midrule
\textit{RuWordNet1.0 - RuWordNet2.0}     & 14 660 & 2 154  \\
\textit{RUSSE'2020}     & 2 288 & 525  \\
\bottomrule
\end{tabular}
\end{table}

\subsection{English Dataset} 

To compile dataset, we choose two versions of WordNet and then select words which appear only in a newer version. For each word, we get its hypernyms from the newer WordNet version and consider them as gold standard hypernyms. We add words to the dataset if only their hypernyms appear in both versions. We do not consider adjectives and adverbs, because they often introduce abstract concepts and are difficult to interpret by context. Besides, the taxonomies for adjectives and adverbs are worse connected than those for nouns and verbs making the task more difficult.

In order to find the most suitable pairs of releases, we compute WordNet statistics (see Table \ref{tab:wn}). New words demonstrate the difference between the current and the previous WordNet version. For example, it shows that the dataset generated by ``subtraction'' of WordNet 2.1 from WordNet 3.0 would be too small, they differ by $678$ nouns and $33$ verbs. Therefore, we create several datasets by skipping one or more WordNet versions.
The statistics for the datasets we select for our study are provided in Table \ref{tab:dataset}.

\begin{table*}[ht]
\centering
\small
\caption{Statistics of the English WordNet taxonomies used in this study.}
\label{tab:wn}
\begin{tabular}{lrrrrrr}
\toprule
\multicolumn{1}{c}{\multirow{2}{*}{\textbf{Taxonomy}}} & \multicolumn{2}{c}{\textbf{Synsets}}                  & \multicolumn{2}{c}{\textbf{Lemmas}}                   & \multicolumn{2}{c}{\textbf{New words}}                      \\ 
\multicolumn{1}{c}{}                                   & \multicolumn{1}{c}{Nouns} & \multicolumn{1}{c}{Verbs} & \multicolumn{1}{c}{Nouns} & \multicolumn{1}{c}{Verbs} &
\multicolumn{1}{c}{Nouns} & \multicolumn{1}{l}{Verbs} \\ \midrule
\textit{WordNet 1.6}                                   &             66 025              &              12 127             &             94 474              &       10 319                    &        -                   &     -    \\
\textit{WordNet 1.7}                                   &             75 804              &              13 214             &             109 195              &       11 088                    &        11 551  &     401        \\
\textit{WordNet 2.0}                                   &             79 689              &              13 508             &             114 648              &       11 306                    &        4 036 & 182 \\
\textit{WordNet 2.1}                                   &             81 426              &              13 650             &             117 097              &       11 488                    &       2 023 & 158  \\
\textit{WordNet 3.0}                                   &           82 115                &           13 767                &          117 798                 &        11 529                   &          678              &   33    \\ 
\bottomrule
\end{tabular}
\end{table*}

\begin{table*}[]
    \centering
     \caption{Examples of Russian nouns with translation mapped to English WordNet.}
    \begin{tabular}{ccccc}
         \toprule
         \textbf{Word}  & \textbf{Translation} & \textbf{RuWordNet hypernyms}                                                             & \textbf{RuWordNet hypernym names}                                                                                                               & \textbf{WordNet hypernyms}                                                                                    \\\midrule

\cyrins{абсентеизм}     & absenteeism          & \begin{tabular}[c]{@{}c@{}}{[}"147309-n", \\ "117765-n", \\ "117017-n"{]}\end{tabular} & \begin{tabular}[c]{@{}c@{}}{[}\cyrins{'неучастие, отказ от участия'}, \\ \cyrins{'уклониться (отказаться)'}, \\ \cyrins{'отказаться, не согласиться'}{]}\end{tabular}    & \begin{tabular}[c]{@{}c@{}}{[}"non-engagement.n.01", \\ "evasion.n.03", \\ "rejection.n.01"{]}\end{tabular} \\ \midrule
\cyrins{кибертерроризм} & cyber terrorism      & \begin{tabular}[c]{@{}c@{}}{[}"7334-n", \\ "4590-n", \\ "2400-n"{]}\end{tabular}       & \begin{tabular}[c]{@{}c@{}}{[}\cyrins{'преступление против} \\ \cyrins{общественной безопасности',} \\ \cyrins{'компьютерное преступление',} \\ \cyrins{'терроризм'}{]}\end{tabular} & \begin{tabular}[c]{@{}c@{}}{[}null, \\ "cybercrime.n.01", \\ "terrorism.n.01"{]}\end{tabular}               \\
\midrule
\cyrins{метропоезд}     & subway train         &
\begin{tabular}[c]{@{}c@{}}{
[}"141975-n", \\
"7133-n"{]}   \end{tabular}         & \begin{tabular}[c]{@{}c@{}}{[}\cyrins{'электрическое} \\ \cyrins{транспортное средство'}, \\ \cyrins{'электропоезд'}{]}\end{tabular} 
& \begin{tabular}[c]{@{}c@{}}{[}null, \\ null{]} \end{tabular}       \\

\bottomrule
    \end{tabular}
   
    \label{tab:ili_example}
\end{table*}

As gold standard hypernyms, we use not only the immediate hypernyms of each \textit{lemma} (initial form of a word --- infinitive for a verb, single number and nominative case for a noun, etc.) but also the second-order hypernyms: hypernyms of the hypernyms. We include them in order to make the evaluation less restricted. According to our empirical observations, the task of automatically identifying the exact hypernym might be too challenging, and finding the ``region'' where a word belongs (``parents'' and ``grandparents'') can already be considered a success.

This method of dataset construction does not use any language-specific or database-specific features, so it could be transferred to other wordnets or taxonomies with timestamped releases.

All datasets\footnote{\url{https://zenodo.org/record/4279821}} created for this research and the code\footnote{\url{https://github.com/skoltech-nlp/diachronic-wordnets}} for their construction are publicly available.

\subsection{Russian Datasets}

In order to create an analogous version to English dataset for Russian, we use the RuWordNet taxonomy~\cite{loukachevitch2016creating}. RuWordNet comprises \textit{synsets} --- sets of synonyms expressing a particular concept. A synset consists of one or more \textit{senses} --- words or multi-word constructions in the initial form. Therefore, we use the current version of RuWordNet and the extended version of RuWordNet which has not been published yet to compile the dataset (cf. Table~\ref{tab:dataset}).

The RUSSE'2020 dataset was created for the Dialogue Evaluation \cite{nikishina2020taxonomy} and can be viewed as a restricted subset of the Russian dataset. In the RUSSE'2020 the following categories of words were excluded:
\begin{itemize}
\item all three-symbol words and the majority of four-symbol words;
\item diminutive word forms and feminine gender-specific job titles; 
\item words which are derived from words which are included in the published RuWordNet;
\item words denoting inhabitants of cities and countries;
\item geographic and personal names;
\item compound words that contain their hypernym as a substring.
\end{itemize} 

\subsection{WordNet ILI mapping (ru-en)}

In order to connect wordnets in different languages the Inter-Lingual Index (ILI) is used \cite{bond-etal-2016-cili}. This mapping is designed to make
possible coordination between wordnet projects. For the Russian test sets we also provide mapping from RuWordNet to WordNet\footnote{\url{https://doi.org/10.5281/zenodo.4969267}}. For each hypernym synset of each query word we present the corresponding WordNet synset index. Table \ref{tab:ili_example} demonstrates several examples of this kind of mapping. As we can see, datasets for the Russian nouns and verbs are extended with additional column called ``WordNet synsets'', where the corresponding WordNet3.0 synsets are listed in accordance with the Russian synset list.

However, not all synsets have an equivalent in the other language, as there exist untranslatable concepts and lacunae. Therefore, we present in the Table \ref{tab:ili} the coverage of the WordNet synsets for the hypernyms of query words from the test set. This mapping can be further used for multilingual experiments.

\begin{table}[h!]
\caption{Coverage of the WordNet synsets for the hypernyms in the Russian test sets.}
\begin{tabular}{lrr}
\toprule
\multicolumn{1}{c}{Input type} & \multicolumn{1}{c}{Total} & \multicolumn{1}{c}{Have ILI mapping} \\
\midrule
\multicolumn{3}{c}{Non-restricted nouns}                                                          \\
\midrule
All synsets                    & 41,694                     & 28,425                                \\
Unique synsets                 & 4,777                      & 2,791                                 \\
Query words                   & 17,475                     & 15,251                                \\
\midrule
\multicolumn{3}{c}{Restricted (private) nouns}                                                    \\\midrule
All synsets                    & 4,456                      & 3,087                                 \\
Unique synsets                 & 1,376                      & 885                                  \\
Query words                   & 1,920                      & 1,720                                 \\\midrule
\multicolumn{3}{c}{non-restricted verbs}                                                          \\\midrule
All synsets                    & 6,783                      & 4,860                                 \\
Unique synsets                 & 1,473                      & 931                                  \\
Query words                   & 2,872                      & 2,606                                 \\\midrule
\multicolumn{3}{c}{restricted (private) verbs}                                                    \\\midrule
All synsets                    & 1,110                      & 821                                  \\
Unique synsets                 & 611                       & 419                                  \\
Query words                   & 477                       & 440               \\
\bottomrule
\end{tabular}

\label{tab:ili}
\end{table}

\subsection{Evaluation Metric}

The goal of diachronic taxonomy enrichment is to build a newer version of a wordnet by attaching the new given terms to the older wordnet version. We cast this task as a soft ranking problem and use Mean Average Precision (MAP) score for the quality  assessment:

\begin{equation}
\begin{array}{c}
\hspace{10mm}MAP = \frac{1}{N} \sum_{i=1}^{N} AP_{i}, \\ \ \\
\hspace{10mm}AP_{i} = \frac{1}{M} \sum_{i}^{n} prec_{i} \times I[y_{i} = 1],
\end{array}
\end{equation}
where $N$ and $M$ are the number of predicted and ground truth values, respectively, $prec_i$ is the fraction of ground truth values in the predictions from 1 to $i$, $y_i$ is the label of the $i$-th answer in the ranked list of predictions, and $I$ is the indicator function.

This metric is widely acknowledged in the Hypernym Discovery shared tasks, where systems are also evaluated over the top candidate hypernyms \cite{camacho-collados-etal-2018-semeval}.
The MAP score takes into account the whole range of possible hypernyms and their rank in the candidate list.

However, the design of our dataset disagrees with MAP metric. As we described in Section \ref{sec:dataset}, the gold-standard hypernym list is extended with second-order hypernyms (parents of parents). This extension can distort MAP. If we consider all gold standard answers as compulsory for the maximum score, it means that we demand models to find both direct and second-order hypernyms. This disagrees with the original motivation of including second-order hypernyms to the gold standard --- it was intended to make the task easier by allowing a model to guess a direct \textit{or} a second-order hypernym.

On the other hand, if we decide that guessing \textit{any} synset from the gold standard yields the maximum MAP score, we will not be able to provide an adequate evaluation for words with multiple direct hypernyms. There exist two cases thereof:

\begin{enumerate}[noitemsep]
    \item the target word has two or more hypernyms which are co-hyponyms or one is a hypernym of the other --- this word has a single sense, but the annotator decided that multiple related hypernyms are needed to reflect all shades of the meaning,
    \item the target word has two or more hypernyms which are not directly connected in the taxonomy and neither are their hypernyms. This happens if:
    \begin{enumerate}[noitemsep]
        \item the word's sense is a composition of senses of its hypernyms, e.g. ``impeccability'' possesses two components of meaning: (``correctness'', ``propriety'') and  (``morality'', ``righteousness'');
        
        \item the word is polysemous and different hypernyms reflect different senses, e.g. ``pop-up'' is a book with three-dimensional pages (``book, publication'') and a baseball term (``fly, hit'').
        
    \end{enumerate}
\end{enumerate}

While the case 2a corresponds to a monosemous word and the case 2b indicates polysemy, this difference does not affect the evaluation process. We propose that in both these cases in order to get the maximum MAP score a model should capture all the unrelated hypernyms which correspond to different components of sense. At the same time, we should bear in mind that guessing a direct hypernym or a second-order hypernym are equally good options. Therefore, following \cite{nikishina2020taxonomy}, we evaluate our models with modified MAP. It transforms a list of gold standard hypernyms into a list of connected components. Each of these components includes hypernyms (both direct and second-order) which form a connected component in a taxonomy graph. (According to graph theory, connected component is a subgraph, in which there is a path between any two nodes.) Thus, in the case 1 we will have a single connected component, and a model should guess \textit{any} hypernym from it to get the maximum MAP score. In the cases 2a and 2b we will have multiple components, and a model should guess \textit{any} hypernym from \textit{each} of the components.

\section{Base Methods}
\label{sec:baselines}

Here we first describe our baseline model which is a method of synset ranking based on distributional embeddings and hand-crafted features (the method was proposed as a baseline for RUSSE-2020 shared task~\cite{nikishina-etal-2020-studying}). We then propose extending it with new features extracted from Wiktionary and use the alternative sources of information about words (e.g. graph representations) and their combinations.

\subsection{Baseline}
\label{baseline}

We consider the approach by Nikishina et al.~\cite{nikishina-etal-2020-studying} as our baseline. There, 
we first create a vector representation for each synset in the taxonomy by averaging vectors (pretrained embeddings) of all words from this synset. 
Then, we retrieve top 10 synsets whose vectors are the closest to that of the \textit{query word} (we refer to these synsets as \textit{synset associates}). For each of these \textit{associates}, we extract their immediate hypernyms and hypernyms of all hypernyms (second-order hypernyms). This list of the first- and second-order hypernyms forms our \textit{candidate set}. We need to rank the candidates by their relevance for the query word. Note that the lists of candidates for different associates can have intersections. When forming the overall candidate set, we make sure that each candidate occurs in it only once.

The intuition behind the method is the following. We propose that if a synset of a taxonomy is a \textit{parent} of a word which is similar to our query word, it can also be a parent of this query word.

To rank the candidate set of synsets we train a Linear Regression model with L2-regularisation on the training dataset formed of the words and synsets of WordNet. Candidate hypernyms are ranked by their model output score. We limit the output to the $k=10$ best candidates.

We rank the candidate set using the following features:


\begin{itemize}
    \item $n \times sim(v_i, v_{h_j})$, where $v_x$ is a vector representation of a word or a synset $x$, $h_j$ is a hypernym, $n$ is the number of occurrences of this hypernym in the merged list, $sim(v_i, v_{h_j})$ is the cosine similarity of the vector of the input word $i$ and hypernym vector $h_j$;
    \item the candidate presence in the Wiktionary hypernyms list for the input word (binary feature);
    \item the candidate presence in the Wiktionary synonyms list (binary feature);
    \item the candidate presence in the Wiktionary definition (binary feature);
    \item the average cosine similarity between the candidate and the Wiktionary hypernyms of the input word.
\end{itemize}



\subsection{DWRank}
\label{sec:DWRank}

We present a new method of taxonomy enrichment --- Distributional Wiktionary-based synset Ranking (\textbf{DWRank}). It combines distributional features with features from Wiktionary. DWRank builds up on the baseline described in Section~\ref{baseline}. We extend the baseline Logistic Regression model with the new features which mainly account for the number of occurrences of a synset in the candidate lists of different synset associates (nearest neighbours) of the query word. We introduce the following new features:
\begin{itemize}
    \item the number of occurrences (\textit{n}) of the synset in the merged candidate list and the quantity $log_2(2 + n)$ which serves for smoothing,
    \item the minimum, average, and maximum proximity level of the synset in the merged candidate list:
    \begin{itemize}
        \item the level is 0 if the synset was added based on similarity to the query word, 
        \item the level of 1 is for the immedidate hypernyms of the query word, 
        \item the level of 2 is for the hypernyms of the hypernyms,
    \end{itemize}
    \item the minimum, average, and maximum similarities of the query word to all words of the synset,
    \item the features based on hyponyms of a candidate synset (``children-of-parents''):
    \begin{itemize}
        \item we extract all hyponyms (``children'') of the candidate synset,
        \item for each word/phrase in each hyponym synset we compute their similarity to the query word,
        \item we compute the minimum, average, and maximum similarity for each hyponym synset,
        \item we form three vectors: a vector of minimums of similarities, average similarities, and maximum similarities of hyponym synsets,
        \item for each of these vectors we compute minimum, average, and maximum. We use these resulting 9 numbers as features.
    \end{itemize}
    These features account for different aspects of similarity of the candidate's children to the query word and help defining if these children can be the query word's co-hyponyms (``siblings'').
\end{itemize}

Moreover, in this approach we use cross-validation and feature scaling when training the Logistic Regression model. 

This methods could be easily extended to other languages that possess a taxonomy, a wiki-based open content dictionary (Wiktionary) and text embeddings like fastText or/and word2vec and GloVe.

\subsection{Web-based Synset Ranking (WBSR)}

In this section, we describe \textbf{WBSR} (Web-based Synset Ranking) a method which 
leverages the power of the existing general-purpose services. It makes use of the two famous search engines: Google (for both English and Russian datasets) and Yandex\footnote{\url{https://yandex.com}} (for Russian only). According to our hypothesis, the search results for a word are likely to contain its hypernyms or co-hyponyms as they are often used to define a word via generalisation or by providing synonyms (co-hyponyms). For instance, if we do not know what ``abdominoplasty'' is, searching for it with a search engine can yield its definition ``a cosmetic surgery procedure''. 

Another source that we could probably benefit from is another taxonomy, preferably larger than the one we work with. However, there might be no other taxonomies available in the same language. Therefore, in this case we can resort to Machine Translation and automatically translate query words into a rich-resource language (e.g. English) in order to use an existing taxonomy (e.g. English  Princeton WordNet). 
In this study we use Yandex Machine Translation system\footnote{\url{https://translate.yandex.ru/}} to translate query words into English and then translate hypernyms (if they are found) back into Russian.

The main drawback of using external sources such as search engines and machine translation systems is their weak reproducibility. The search results are dependent on the search history, so reproducing the experiment on a different account or after a relatively long period of time is problematic.
However, since the method greatly improves the performance even with trivial handling of the collected data, we use it despite its drawback.
To make our results reproducible, we release all data from the external sources used in our approach.\footnote{\url{https://doi.org/10.5281/zenodo.4540717}}

Similarly to the approaches described above, here we also make use of Wiktionary and fastText embeddings cosine similarity. However, we treat synsets and words/phrases that they consist of in a different way. 
In the previously described approaches we computed embeddings for multiword phrases by averaging word embeddings of individual words in them. Here we treat them as sentences --- we compute their embeddings using the \texttt{get\_sentence\_vector} method from fastText Python library. There, fastText vectors are divided by their norms and then averaged, so that only vectors with the positive $L_2$-norm value are considered. Secondly, we do not combine the word/phrase vectors into a synset vector but operate with the word/phrase embeddings directly. 


Similarly to DWRank, the algorithm consists of two steps: candidate generation and candidate ranking. Our candidate list is formed of the following synsets:

\begin{itemize}[noitemsep]
  \item synsets which contain words/phrases from the list of top-10 nearest neighbours of the query word; 
  \item hypernyms and second-order hypernyms of those synsets; 
  \item Wiktionary-based candidates:
  \begin{itemize}
      \item synsets that contain words/phrases listed in Wiktionary as the hypernyms of the query word; 
      \item hypernym synsets of these synsets;
  \end{itemize}
   \item cross-lingual candidates (for the Russian language only):
   \begin{itemize}
        \item synsets that contain words/phrases listed in the English WordNet as the hypernyms of the query word; 
      \item hypernym synsets of these synsets.
   \end{itemize}
 
\end{itemize}

Analogously to DWRank, we then rank all candidates by a logistic regression model which uses the following features:
\begin{itemize}[noitemsep]
  \item the candidate synset contains a word/phrase from the list of query word's nearest neighbours;
  \item the candidate synset is a hypernym of one of the nearest neighbours;
  \item the candidate is a second-order hypernym of one of the nearest neighbours;
  \item the candidate synset contains a hypernym of the query word from Wiktionary;
  \item the candidate synset is a hypernym of the synset which contains a hypernym of the query word from Wiktionary;
  \item the candidate synset contains a word which is present in the definition of the query word from Wiktionary;
  \item the candidate synset is present in the list of English WordNet synset candidates (for the Russian language only);
  \item the candidate synset is present in the list of hypernyms of the WordNet candidates (for the Russian language only);
  \item the candidate synset contains words which occur on the Google results page;
  \item the candidate synset contains words which occur on the Yandex results page (for the Russian language only).
\end{itemize}

The training set for the logistic regression model is formed from a wordnet in the relevant language as follows. For each query word in the list of query words we first find the most similar lemma which is contained in the wordnet. We know hypernyms for these lemmas and use them to generate the training set. We generate the candidate list as described before. First- and second-order hypernyms in this candidate list are used as positive examples for the corresponding lemmas, and synsets from the candidate list which are not hypernyms are considered negative examples.





This approach participated in the RUSSE'2020 Taxonomy Enrichment task for the Russian Language. The method achieved the best result on the nouns track. Therefore, we consider it as the-state-of-the-art method for Russian.

\subsection{WordNet Path Prediction}

A completely different approach to make use of fastText embeddings is presented in the work of Cho et al.~\cite{cho-etal-2020-leveraging}. The authors experiment with encoder-decoder models in order to solve the task of the direct hypernym prediction. They use a standard LSTM-based sequence-to-sequence model~\cite{sutskever2014sequence} with Luong attention~\cite{luong-etal-2015-effective}. First, they average fastText embeddings for the input word or phrase and put it through the encoder. The decoder sequentially generates a chain of synsets from the encoder hidden state, conditioned on the previously generated ones. The authors consider two different setups:
\begin{itemize}
    \item \textit{hypo2path} --- given the input word, generate a sequence of synsets starting from the root synset and going down the taxonomy to the closest hypernym;
    \item \textit{hypo2path reverse} --- given the input word, generate a sequence of synsets starting from the closest hypernym up to the root entity.
\end{itemize}

To be able to apply this sequence-to-sequence architecture to our data, we build new datasets similar to the ones described in~\cite{cho-etal-2020-leveraging}. We generate a path from the WordNet starting from the root node to the target synset or word. Analogously to the original work, we include multiple paths from the root to the parents of the query word. We filter the validation set to only include queries that do not occur anywhere in the full taxonomy paths of the training data. To sort candidates generated by the decoder, we enumerate the generated \textit{hypo2path} sequence from the right to the left or the \textit{hypo2path reverse} from the left to the right and get the first 10 synsets. 

Additionally, we extend this approach by replacing the LSTM+attention architecture with the Transformer architecture~\cite{NIPS2017_3f5ee243}. During training we provide an embedding of a synset as input to the Transformer and expect the model to generate a sequence of synsets starting from the hypernym of the input synset. During inference we provide embedding of query words as input  
expect the model to output sequences of synsets starting with the direct hypernyms.

\subsection{Word Representations for DWRank}

We test our baseline approach and DWRank with different types of embeddings: fastText~\cite{bojanowski-etal-2017-enriching}, word2vec~\cite{word2vec} embeddings for English and Russian datasets and also GloVe embeddings~\cite{pennington-etal-2014-glove} for the English dataset.

We use the fastText embeddings from the official website\footnote{\url{https://fasttext.cc/docs/en/crawl-vectors.html}} for both English and Russian, trained on Common Crawl from 2019 and Wikipedia CC including lexicon from the previous periods as well. For word2vec we use models from~\cite{fares-etal-2017-word,KutuzovKuzmenko2017} for both English\footnote{\url{http://vectors.nlpl.eu/repository/20/29.zip}} and Russian.\footnote{\url{http://vectors.nlpl.eu/repository/20/185.zip}}
We lemmatise words and synsets for both languages with the same UDPipe~\cite{udpipe:2017} model which was used while training the representations. 
For the out-of-vocabulary (OOV) words we find all words in the vocabulary with the longest prefix matching this word and average their embeddings like in~\cite{dale2020russe}. As for the GloVe embeddings, we also use them from the official website\footnote{\url{https://nlp.stanford.edu/projects/glove/}} trained on Common Crawl, the vocabulary size is 840 billion tokens.

\section{DWRank-Graph}
\label{sec:graph}


The DWRank method extracts a set of candidate synsets based on the similarities of word vectors. So far we used only distributional word vectors (fastText, GloVe, etc.) to represent words. On the other hand, graph-based representations can contain the taxonomic information which is absent in distributional embeddings~\cite{makarov2021fusion}. 

Here, we present \textbf{DWRank-Graph}. This is the same DWRank method where the distributional embeddings are replaced with graph representations. 
The score prediction model and the features it uses do not change. Below we describe the graph representations and their combinations we applied in DWRank-graph.

\subsection{Poincaré Embeddings}

Poincaré embeddings is an approach for ``learning hierarchical representations
of symbolic data by embedding them into hyperbolic space --- or more precisely into
an ''$n$-dimensional Poincaré ball''~\cite{nickel2017poincare}. Poincaré models are trained on hierarchical structures and simultaneously capture hierarchy and similarity due to the underlying hyperbolic geometry. 
According to the authors, hyperbolic embeddings are more efficient on the hierarchically structured data and may outperform Euclidean embeddings in several tasks, e.g, in Taxonomy Induction \cite{aly-etal-2019-every}. 

Therefore, we use Poincaré embeddings of our wordnets for the taxonomy enrichment task. We train Poincaré ball model for our wordnets using the default parameters and the dimensionality of 10, which yields the best results on the link prediction task \cite{nickel2017poincare}.

However, applying these embeddings to the task is not straightforward, because
Poincaré model's vocabulary is non-extensible. It means that new words that we need to attach to the existing taxonomy will not have any Poincaré embeddings at all and we cannot make use of the embeddings similarity. To overcome this limitation, we compute top-5 fastText nearest synsets (analogously to the procedure described in Section \ref{baseline}) 
and then aggregate embeddings in hyperbolic space using Einstein midpoint, following~\cite{guh2016hyperbolic}. The resulting vector is considered as an embedding of the input word in the Poincaré space.

Then, we use vectors from this vector space to generate candidates for the DWRank approach. As the model we present in Section~\ref{sec:DWRank} does not depend on the types of input embeddings, we are able to provide Poincaré embeddings as input.

\subsection{Node2vec Embeddings}

The hierarchical structure of the taxonomy is 
a graph structure, and we may also consider taxonomies as  graphs and apply random walk approaches to compute embeddings for the synsets. For this purpose we apply node2vec~\cite{node2vec-kdd2016} approach which represents a ``random walk of a fixed length $l$'' with ``two parameters $p$ and $q$ which guide the walk in breadth or in depth''. 
Node2vec randomly samples sequences of nodes and then applies the skip-gram model~\cite{DBLP:journals/corr/abs-1301-3781} to train their vector representations. 
We train node2vec representations of all synsets in our wordnets with the following parameters: $dimensions=300$, $walk\_length=30$, $num\_walks=200$. The other parameters are taken from the original implementation. 

However, analogously to Poincaré vector space, 
node2vec model has no techniques for representing out-of-vocabulary words. Thus, it is unable to map new words to the vector space. 
To overcome this limitation, we apply the same technique of averaging top-5 nearest neighbours from fastText and considering their mean vector as the new word embedding and search for the most similar synsets.


\subsection{Graph Neural Networks}

The models described above have a major shortcoming:  the resulting vectors for the input words heavily depend on their representations in the fastText model. This can lead to the incorrect results if the word's nearest neighbour list is noisy and does not reflect its meaning. In this case the noise will propagate through the graph embedding (Poincaré or node2vec) model and result in inaccurate output even if the graph embedding model is of high quality.

Therefore, we test different graph neural network (GNN) architectures --- Graph Convolutional Network~\cite{kipf2016semi}, Graph Attention Network \cite{velickovic2018graph} and GraphSAGE \cite{hamilton2017inductive}  (SAmple and aggreGatE) which make use of both fastText embeddings and the graph structure of the taxonomy. 

All the above mentioned models work similarly. According to \cite{makarov2021fusion}, ``GCN works similarly to the
fully-connected layers for neural networks. It multiplies weight matrices with the original features but
masking them with an adjacency matrix. Such a method allows to account not only node representation, but also representations of neighbours and two-hop neighbours.'' The GraphSAGE model addresses the problem of unseen nodes representation 
by training a set of aggregator functions that learn to aggregate feature information from a node’s local neighborhood. GCN, on the contrary, learns a distinct embedding vector for each node. In GAT, the convolution operation from GCN is replaced with the attention mechanism. It uses the self-attention mechanism of Transformers~\cite{transformer} to aggregate the information from the one-hop neighbourhood.

FastText embeddings are used as input node features for all models, which is definitely an advantage of the model over Poincaré and node2vec, as they do not use word embeddings for training. Even though new words are not connected to the taxonomy, it is still possible to compute their embeddings according to their input node features. 

We get the vector representations of query words from one of the pre-trained GNN models and then use them as the input to DWRank. Even though all methods work similarly, they demonstrate different performance on different datasets.

\subsection{Text-Associated Deep Walk}

Text-Associated Deep Walk (TADW)~\cite{yang2015network} is another approach that incorporates text and graph information into one vector representation. The method is based on the DeepWalk algorithm~\cite{perozzi2014deepwalk} which learns feature representations by simulating uniform random walks. To be specific, the sampling strategy in DeepWalk can be seen as a special case of node2vec with $p=1$ and $q=1$. 

The authors prove that the DeepWalk approach is equivalent to matrix factorization. They incorporate text features of vertices into network representation learning within the framework of matrix factorization. First, they define
matrix $M \in \mathbb{R}^{|V|\times|V|}$ where each entry $M_{ij}$ is the logarithm of the average probability that vertex $v_i$ randomly walks to vertex $v_j$ in a fixed number of steps. In comparison to DeepWalk, where the goal is to factorize matrix $M$ into the product of two low-dimensional matrices
$W \in \mathbb{R}^{k\times |V|}$
and $H \in \mathbb{R}^{k \times |V|}$ ($k \ll |V|$), TADW aims to factorize matrix $M$ into
the product of three matrices: $W \in \mathbb{R}^{k\times|V|}$, $H \in \mathbb{R}^{k\times f_t}$ and
text features $T \in \mathbb{R}^{f_t \times |V|}$. As text features, in this work we apply fastText embeddings.

After having learnt the factorisation of matrix $M$, we use rows of matrix $W$ as node (synset) embeddings in DWRank.

\subsection{High-Order Proximity preserved Embeddings}

High-Order Proximity preserved Embeddings (HOPE) \cite{ou2016asymmetric} is yet another approach that embeds a graph into a vector space preserving information about graph properties and structure. Unfortunately, most structures cannot preserve the asymmetric transitivity, which is a critical property of directed graphs. To solve the problem, the authors employ matrix factorization to directly reconstruct asymmetric distance measures like Katz index, Adamic-Adar or common neighbors. 
This approach is scalable --- it preserves high-order
proximities of large scale graphs, and capable of capturing
the asymmetric transitivity. HOPE outperforms state-of-the-art algorithms in tasks of reconstruction, link prediction and vertex recommendation.

As for our Taxonomy Enrichment task, we also apply HOPE to generate graph embeddings to be used as input in the DWRank model. The main difference with the other embeddings is that HOPE does not incorporate textual information from the nodes.

\section{DWRank-Meta} 
\label{sec:dwrank_meta}

In DWRank we employed only distributional information, i.e. pre-trained word embeddings, whereas in DWRank-Graph we represented words using the information from the graph structure of the taxonomy and usually ignoring their distributional properties. Meanwhile, taxonomy enrichment models may benefit from combining these two types of information. Therefore, we present \textbf{DWRank-Meta} --- an extension of DWRank which combines multiple types of input word representations. 

As in DWRank-Graph, the process of candidates selection, the feature set and the algorithm of synset ranking stay intact. Here we change the input representations of words and synsets.

\subsection{Base Meta-Embeddings}

The easiest way of combining embeddings of different types is to concatenate them and use the concatenated vector as an input. We refer to this method as \textbf{Concat} and combine different subsets of distributional (fastText, word2vec, GloVe) and graph-based (Poincaré, node2vec, GCN, GAT, GraphSAGE, TADW, HOPE) embeddings. In addition to that, we perform Singular value decomposition (SVD) over this concatenation as proposed in~\cite{yin2016learning}. This approach is referred to as \textbf{SVD}.

\subsection{Autoencoded Meta-Embeddings}

We propose using two variants of autoencoders for the generation of meta-embeddings: Concatenated Autoencoded Meta-Embeddings (CAEME) and Averaged Autoencoded Meta-Embeddings (AAEME)~\cite{bollegala2018learning}. They have shown good results on the task of evaluating lexical similarity. However, they have never been applied to taxonomy enrichment.

We generate meta-embeddings as follows. Let us consider an embedding model $s(w)$. 
For each of such embedding models we train an autoencoder consisting of an encoder and a decoder:




\begin{equation}
\begin{array}{c}
\hspace{13mm}E(s(w)) = h(w),\\ \\ \hspace{13mm}D(h(w)) = \hat{s}(w),\\ \\ \hspace{13mm}L_{ED} = dist(s(w), \hat{s}(w)),
\end{array}
\end{equation}

where $E$ and $D$ are the encoder and the decoder, and $L$ is the loss used for training of the autoencoder. The loss is implemented as the distance (\textit{dist}) between the original and the reconstructed embeddings. The \textit{dist} can be any distance or similarity measure such as MSE, KL-divergence, or cosine distance. In our preliminary study, the cosine distance showed the best results, so we use it in our experiments.


Let us consider two embedding models $s_1(w)$ and $s_2(w)$. In such a case, the input to each decoder is not the result of the corresponding encoder, but meta-embeddings, which depends on the both encoders. Depending on the approach, meta-embeddings can be built in different ways, we construct the meta-embeddings as follows. In case of CAEME, we take an $L_2$-normalised concatenation of the two source embeddings encoded with respective encoders $E_1(s_1(w))$ and $E_2(s_2(w))$:

\begin{equation}
m(w)=\frac{E_1(s_1(w))\oplus E_2(s_2(w))}{||E_1(s_1(w))\oplus E_2(s_2(w)||_2},
\label{eq:caeme}
\end{equation}

where $\oplus$ is the concatenation operation.

The drawback of this model is the growing dimensionality of meta-embeddings for cases where we combine multiple source embeddings. To fight that, we can replace the concatenation operation with averaging, yielding AAEME. It computes meta-embedding of a word $w$ from its two source embeddings $s_1(w)$ and $s_2(w)$ as the $L_2$-normalised sum of internal representations $E_1(s_1(w))$ and $E_2(s_2(w))$:

\begin{equation}
m(w)={{E_1 (s_1(w))+E_2(s_2(w))}\over{||E_1 (s_1(w))+E_2(s_2(w))||_2}}. 
\end{equation}

In CAEME, the dimensionality of the meta-embedding space is the sum of the dimensions of the source embeddings, whereas in AAEME it stays the same. The AAEME encoder can be seen as a special case of the CAEME encoder where  the meta-embedding is computed by averaging
the two encoded sources in equation~\ref{eq:caeme} instead of their concatenation. 








\subsection{Training of Autoencoders}

We can impose additional restrictions on *AEME models during training. One of such restrictions is the use of triplet loss. 
We restrict a word to be closer to the words that are semantically related to it according to the taxonomy than to a randomly chosen word with some $margin$: 

\begin{equation}
\begin{array}{c}
L(w_a, w_p, w_n) = max(||m(w_a) - m(w_p))|| - \\||m(w_a) - m(w_n))|| + margin, 0),
\end{array}
\label{eq:triplet_loss}
\end{equation}

where ||.|| is a distance function, $w_a$ is the target word, $w_p$ and $w_n$ are positive and negative words, respectively.

The algorithm of calculating triplet loss is as follows:

\begin{enumerate}
    \item for each word presented in the taxonomy, we compile a list of semantically related words which includes synonyms, hyponyms and hypernyms;
    \item at each epoch, we randomly select $K$ positive words from this related words set and form a set of $K$ negative words by selecting them randomly from the vocabulary;
    \item if the word is not presented in the taxonomy, then we cannot form a list of related words for it. In this case, we generate positive vectors for it by adding random noise to its vector;
    \item next, we calculate the triplet margin loss by combining the triplet loss with the original loss as $\alpha * loss + (1 - \alpha) * triplet\_loss$.
\end{enumerate}

We use the following parameters for the triplet loss: $K = 5$, $margin = 0.1$, $alpha = 0.005$. These parameters were selected via grid search with AAEME algorithm on the English 1.7 dataset.

\section{Experiments}
\label{sec:evaluation}
In this section, we report and discuss the performance of our models in the taxonomy enrichment task. 
We experiment with our DWRank approach and its modifications DWRank-Graph and DWRank-Meta. In addition to that, we compare them with the baseline introduced in RUSSE'2020 shared task and with a number of state-of-the-art methods. We conduct the experiments with English and Russian wordnets.

\begin{figure*}[h]
    \centering
    \caption{Comparison of the method performance on \textit{nouns\_1.6} dataset for English. Each colour denotes the method type and the embeddings type used.}
    \label{fig:all_en_nouns}
    \includegraphics[width=\textwidth]{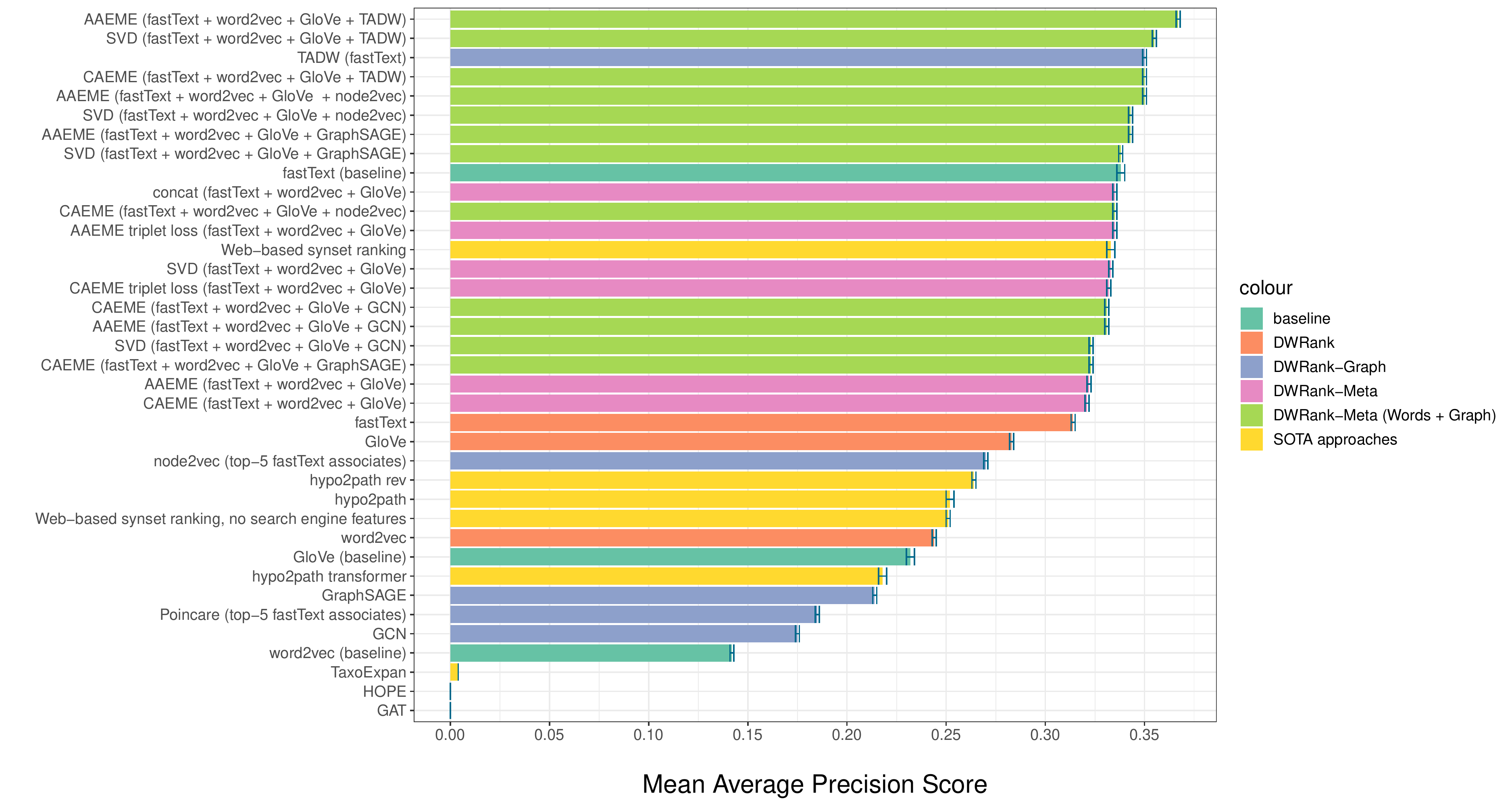}
\end{figure*}

\subsection{Experimental Setup}

For each result we add the standard deviation values. We calculate them as follows. We randomly sample 80\% of the test data and calculate the MAP scores on that part of the test set. We repeat the same procedure 30 times and then calculate the standard deviation on those 30 MAP values.

The MAP metric should be interpreted as follows: the higher the score, the better the results. For instance, $MAP@k = 1.0$ means that all of the $N$ correct hypernyms are present in the first top-$N$ positions in the ranked list of candidates. Moreover, with $MAP@k = 1.0$ the first candidate is always correct. 
$MAP@k \geq 0.5$ means that at least one of the correct hypernyms is present in the two first positions (\textit{top-2}) in the ranked lists of candidates. 
$MAP@k \geq 0.3$ means that at least one of the correct hypernyms is present in the first three positions (\textit{top-3}) in the ranked lists of candidates.

We show the performance of different methods on the nouns attribution task for English (nouns 1.6) in Figure~\ref{fig:all_en_nouns} and for Russian (non-restricted nouns) in Figure~\ref{fig:all_ru_nouns}. The X axis shows the MAP score for each method, the methods are listed along the Y axis. For DWRank-Meta models, we list the embeddings used in the model in brackets. 
The colors of bars in figures correspond to different types of input embeddings. The orange color stands for the vanilla DWRank -- DWRank which uses only distributional embeddings. Purple denotes the  DWRank-Graph variants. For the DWRank-Meta approaches exploiting only distributional embeddings we use pink, and DWRank-Meta on word and graph embeddings is denoted with the bright green color. Previous SOTA approaches are shown in yellow.

The full results for all experiments on the English and Russian datasets can be seen in Appendix~\ref{appendix} in Tables~\ref{tab:all_en} and~\ref{tab:all_rus}, respectively. Here, when listing embeddings used in DWRank-Meta models, we use the shortcut ``\textit{words}'' to denote the combination of fastText, word2vec, and GloVe embeddings.

\subsection{Results}

\begin{figure*}[ht]
    \centering
    \caption{Performance of different models on the Russian non-restricted dataset. Each colour denotes the method type and the embeddings type used.}
    \label{fig:all_ru_nouns}
    \includegraphics[width=\textwidth]{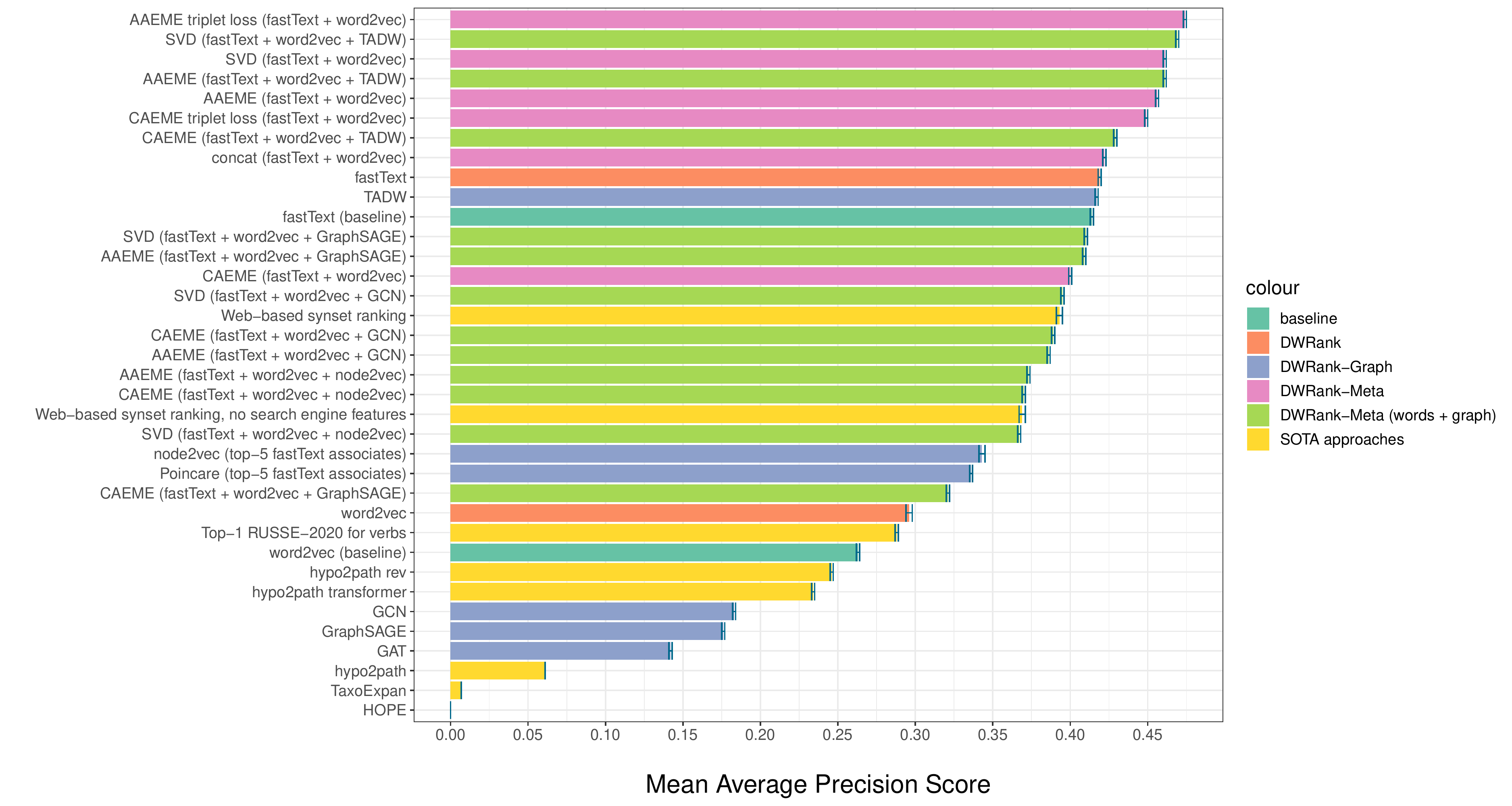}
\end{figure*}

\paragraph{\bf DWRank-Meta} Figures~\ref{fig:all_en_nouns} and~\ref{fig:all_ru_nouns} show that the leaderboard for both English and Russian nouns is dominated by DWRank-Meta models. While English benefits from the union of distributional and graph embeddings, for Russian distributional embeddings alone perform on par with their combinations with graph embeddings. Besides that, high-performing variants of DWRank-Meta for English feature TADW, node2vec, and GraphSAGE, whereas for Russian TADW is the only graph embedding model which does not decrease the scores of DWRank-Meta. 

We see that triplet loss significantly improves the results for DWRank-Meta models (cf. AAEME/CAEME with and without triplet loss) for both English and Russian.

\paragraph{\bf DWRank-Graph} On the other hand, DWRank-Graph fails in the task of taxonomy extension for all datasets. TADW model is the only graph embedding model which can compete with DWRank-Meta models. This can be explained by the fact that TADW is an extended version of DeepWalk and applies the skip-gram model with the pre-trained fastText representations. In contrast to that, the other graph models suffer from the noisy representations of OOV query words.

At the same time, despite the success of TADW, it does not outperform models based solely on distributional embeddings, showing that graph representations apparently do not contribute any information which is not already contained in distributional word vectors.

\paragraph{\bf Baselines} We also notice that for both languages the baselines are quite competitive. They are substantially worse than the best-performing models, but they are much simpler to implement and is fast and easy to train. Therefore, we suggest that it should be preferred in the situation of limited resources and time. 

However, the choice of embedding models is crucial for the baselines (as well as for the vanilla DWRank which performs closely). We see that fastText outperforms word2vec and GloVe embeddings for almost all languages and datasets. The low scores of GloVe and word2vec embeddings on baseline and DWRank methods can be explained by data coverage issues. Fixed vocabularies of word2vec and Glove do not allow generating any representation for missing query words, whereas fastText can handle them.

\paragraph{\bf SOTA models} Neither of SOTA models managed to outperform the fastText baseline or approach the best DWRank-Meta variants. Web-based synset ranking (WBSR) model shows that the information from online search engines and Machine Translation models is beneficial for the task -- its performance without this information drops dramatically. However, this information is not enough to outperform the word embedding-based models.

The performance of hypo2path model is even lower than that of WBSR. Being an autoregressive generative model, it is very sensitive to its own mistakes. Generating one senseless hypernym can ruin all the following chain. Conversely, when starting with the root hypernym ``entity.n.01'', it often takes a wrong path. 
Finally, TaxoExpan model relies on definitions of words which we did not provide in this task. Therefore, its results are close to zero. We do not consider them credible and provide them in italics.

\paragraph{\bf Performance for different datasets}  Figures~\ref{fig:all_en_nouns} and~\ref{fig:all_ru_nouns} as well as the results in the appendix show that there is no single best-performing model. While DWRank-Meta is almost always the best, for different datasets different variations of this model are the most successful. The results are usually consistent for the same language and part of speech (e.g. for different versions of English nouns datasets the best-performing model is the same), but there are exceptions to this regularity. 

\paragraph{\bf Interpretable Evaluation}

MAP metric which is the standard way of evaluating taxonomy enrichment models has a serious drawback. Namely, it is not interpretable, which hampers the understanding of the models' performance. 

Therefore, in addition to MAP we report the Precision@k score which can be interpreted as the ratio of correct synsets in the \textit{top-k} outputs of a model. We evaluate our best systems automatically with the Precision@k ($k=1, 2, 3$) score. The choice of the values of $k$ is explained by the fact that the average number of true ancestors is 2 for the English words and 3 for the Russian words. Thus, Precision@k for $k>3$ will be unfairly understated, because there will always be at most 3 correct answers out of k. This means that for the $k=4$ the maximum is 0.75, for $k=5$ it is 0.6, etc.

\begin{table}[]
\begin{tabular}{l|r|r|r} \toprule
          \textbf{Method}  & \multicolumn{1}{c|}{\begin{tabular}[c]{@{}c@{}}\textbf{Pr@1}\end{tabular}} &\multicolumn{1}{c|}{\begin{tabular}[c]{@{}c@{}}\textbf{Pr@2}\end{tabular}} & \multicolumn{1}{|c}{\begin{tabular}[c]{@{}c@{}}\textbf{Pr@3}\end{tabular}} \\ \midrule
          \multicolumn{4}{c}{English nouns 1.6-3.0} \\ \midrule
 baseline &                  0.260 & 0.184                                                                                     &     0.144                                                                                               \\
 DWRank &            0.245 & 0.170                                                                                           & 0.135                                                                                                    \\
DWRank-Meta &         0.288 & 0.185                                                                                            &  0.143                                                                                                 \\
 DWRank-Graph &                     0.278                                                                       & 0.189           &                       0.150                                                                             \\ 
  DWRank-Meta (Words + Graph) &      \textbf{0.311} & \textbf{0.199}                                                                                                 &               \textbf{0.154}                                                                                   \\ \midrule
 \multicolumn{4}{c}{English verbs 1.6-3.0} \\ \midrule
 baseline  & 0.173 & 0.126                                                                                             &    0.101                                       \\
 DWRank &            \textbf{0.260} & \textbf{0.169}                                                                                           &     0.126                                                                                               \\
DWRank-Meta &          0.238 & 0.168                                                                                             & \textbf{0.131}                                                                                                   \\
 DWRank-Graph &          0.238 & 0.158                                                                                             &  0.118                                                                                                  \\ 
  DWRank-Meta (Words + Graph) &    0.259 & 0.164                                                                                                   &                  0.124                                                                                  \\
   \midrule
 \multicolumn{4}{c}{Russian non-restricted nouns} \\ \midrule
 baseline &   0.346 &          0.228                                                                                    &     0.171                                                                                                         \\
 DWRank &                   0.347                                                                                    &  0.228 & 0.172                                                                                                  \\
DWRank-Meta &           0.396 & \textbf{0.257}                                                                                           &  \textbf{0.196}                                                                                                  \\
 DWRank-Graph &                   0.347                                                                                    &       0.224 &                   0.168                                                                          \\ 
  DWRank-Meta (Words + Graph) & \textbf{0.397} & 0.255                                                                                                      &         0.192                                                                                           \\ \midrule
 \multicolumn{4}{c}{Russian non-restricted verbs}\\ \midrule
   baseline &     0.251      &       0.181                                                                                       &        0.139                                                                                         \\
 DWRank &                0.282 &      0.196                                                                                 &                       0.154                                                                             \\
DWRank-Meta &                \textbf{0.368}          &   \textbf{0.245                      }                                                      &    \textbf{0.190 }                                                                                                 \\
 DWRank-Graph &                  0.274                                                                           & 0.191          &     0.149                                                                                               \\ 
  DWRank-Meta (Words + Graph) &    0.341 & 0.231                                                                                                   &                       0.180                                                                             \\
\bottomrule
\end{tabular}
\caption{Precision@k for the best-performing models for the English and Russian nouns datasets.}
\label{tab:precision_k}
\end{table}

Table~\ref{tab:precision_k} shows the Precision@k scores for the best performing English and Russian models on the nouns datasets. Both of them are DWRank-Meta models with AAEME autoencoders. The English model uses three types of distributional embeddings and TADW graph embeddings, while the Russian model uses only fastText and word2vec but benefits from the triplet loss. We see that Precision@k is particularly high for Russian. There, over a half of generated lists contain a correct synset in the first position. This shows that DWRank-Meta can successfully be used as a helper tool for taxonomy extension.

The results for English are lower. However, this should not be considered as a sign of lower performance of models for English. The Russian and English datasets consist of different words, so they cannot be directly compared.


\section{Error Analysis}
\label{sec:error_analysis}

To better understand the difference in systems performance and their main difficulties, we made a quantitative and qualitative analysis of the results.

\begin{figure*}[ht!]
    \centering
    \caption{Distribution of words over the number of senses.}
    \label{fig:distributions}
    \subfloat[Russian dataset (nouns and verbs)]{{\includegraphics[width=0.5\textwidth]{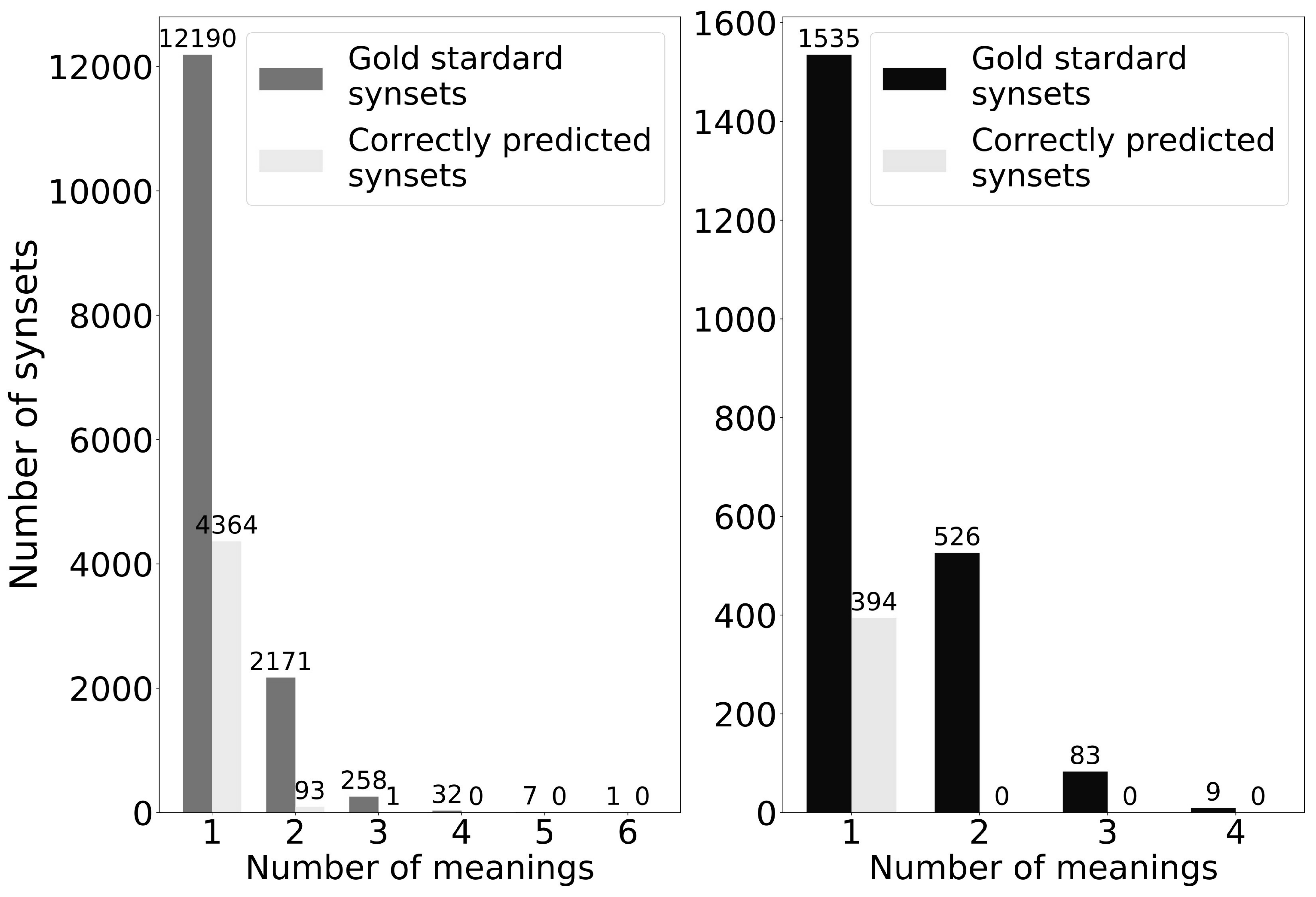} }}%
    \subfloat[English dataset (nouns and verbs)]{{\includegraphics[width=0.5\textwidth]{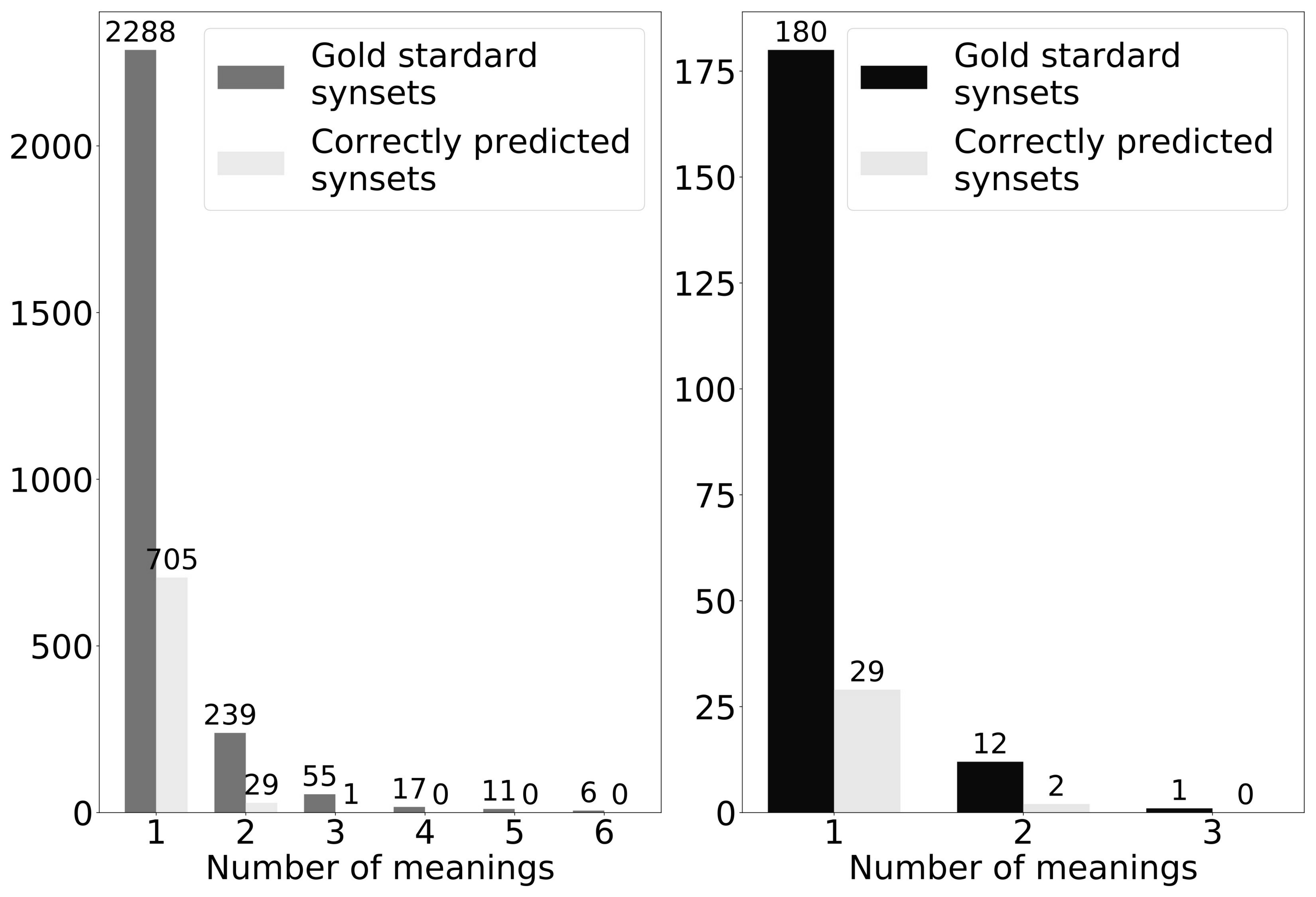} }}%
    
\end{figure*}

\subsection{Comparison of Graph-based Approaches with Word Vector Baselines}

First of all, we wanted to know to what extent 
the set of correct answers of graph-based models overlaps with the one of fastText-based models. In other words, we would like to know if the graph representations are able to discover hypernymy relations which could not be identified by word embeddings.

Therefore, for each new word we computed average precision ($AP$) score and compared those scores across different approaches. 
We found that at least 90\% words for which fastText failed to identify correct hypernyms (i.e. words with $AP=0$) also have the $AP$ of 0 in all the graph-based models. This means that if fastText cannot provide correct hypernyms for a word, other models cannot help either. Moreover, all words which are correctly predicted by graph-based approaches, are also correctly predicted by fastText. 
Moreover, only 8\% to 55\% words correctly predicted by fastText are also correctly predicted by any of the graph-based models. At the same time, the number of cases where graph-based models perform better than fastText is very low (3--5\% cases). 
Thus, combining them cannot improve the performance significantly. This observation is corroborated by the scores of the combined models. 

To contrast the performance of the text and graph embeddings and to demonstrate the input and the output formats of the models we present Tables \ref{tab:examples_en} and \ref{tab:examples_en2} in Appendix \ref{appendix}.
They demonstrate the main features of the tested approaches. The examples do not pretend do be the general case example, however, they do provide the idea about ranking of the results and the performance of text, graph and fusion embedding types.

\subsection{Performance on Polysemous Words}




The differences in word semantics make the dataset uneven. In addition to that, we would also like to understand whether the performance of models depends on the number of connected components (possible meanings) for each word. Thus, we examine how many words with more than one meaning can be predicted by the system.

Figure \ref{fig:distributions}
depicts the distribution of synsets over the number of senses they convey. As we can see, the vast majority of words are monosemous. For Russian nouns, the system correctly identifies almost half of them, whereas for other datasets the share of correctly predicted monosemous words is below 30\%. This stems from the fact that for distributional models it is difficult to capture multiple senses in one vector. They usually capture the most widespread sense of a word. Therefore, the number of predicted synsets with two or more senses is extremely low. A similar power law distribution would be obtained using BERT embeddings, as we are still averaging embeddings from all contexts. This may be one of the reasons why contextualised models did not perform better than the fastText models which capture the main meaning only but do it well.

\subsection{Error Types}

In order to understand why a large number of word hypernyms (at least 60\%) are too difficult for models to predict, we turn to manual analysis of the system outputs. We find out that errors can be divided into two groups: system errors caused by distributional models limitations and taxonomy inaccuracies.
Therefore, we come across five main error types:

{\textbf{Type 1.} Extracted nearest neighbours can be semantically related words but not necessary co-hyponyms:}
    \begin{itemize}[noitemsep]
        \item delist (WordNet); expected senses: get rid of; predicted senses: remove, delete;
        \item \cyrins{хэштег} (hashtag, RuWordNet); expected senses: \cyrins{отличительный знак, пометка} (tag, label); predicted senses: \cyrins{символ, короткий текст} (symbol, short text).
    \end{itemize}
    
{\textbf{Type 2.} Distributional models are unable to predict multiple senses for one word:}
    \begin{itemize}[noitemsep]
        \item latakia (WordNet); expected senses: tobacco; municipality city; port, geographical point; predicted senses: tobacco;
        \item \cyrins{запорожец} (zaporozhets, RuWordNet); expected senses: \cyrins{житель города} (citizen, resident); \cyrins{марка автомобиля, автомобиль} (car brand, car); predicted senses: \cyrins{автомобиль, мототранспортное средство, марка автомобиля} (car, motor car, car brand).
    \end{itemize}
    
\textbf{Type 3.} System predicts too broad / too narrow concepts:
    \begin{itemize}[noitemsep]
        \item midweek (WordNet); expected senses: day of the week, weekday; predicted senses: time period, week, day, season;
        
        
        \item \cyrins{медянка} (smooth snake, RuWordNet); expected senses: \cyrins{неядовитая змея, уж} (non-venomous snake, grass snake); predicted senses: \cyrins{змея, рептилия, животное} (snake, reptile, animal).
        

    \end{itemize}
    
\textbf{Type 4.} Incorrect word vector representation: nearest neighbours are semantically far from the meaning of the inputt word:
    \begin{itemize}[noitemsep]
        \item falanga (WordNet); expected senses: persecution, torture; predicted senses: fish, bean, tree, wood.;
        
        \item \cyrins{кубокилометр} (cubic kilometer, RuWordNet); expected senses: \cyrins{единица объема, единица измерения} (unit of capacity, unit of measurement); predicted senses: \cyrins{город, городское поселение, кубковое соревнование, спортивное соревнование} (city, settlement, competition, sports contest).

    \end{itemize}
    
\textbf{Type 5.} Unaccounted senses in the gold standard datasets, inaccuracies in the manual annotation:
    \begin{itemize}[noitemsep]
        \item emeritus (WordNet); expected senses: retiree, non-worker; predicted senses: professor, academician;
        \item \cyrins{сепия} (sepia, RuWordNet); expected senses: \cyrins{морской моллюск} ``sea mollusc''; predicted senses: \cyrins{цвет, краситель} (color, dye).

    \end{itemize}

\begin{figure*}%
    \centering
    \caption{Manual datasets evaluation results: Precision@10.}%
    \label{fig:precision}%
    \subfloat[Russian dataset]{{\includegraphics[width=0.5\textwidth]{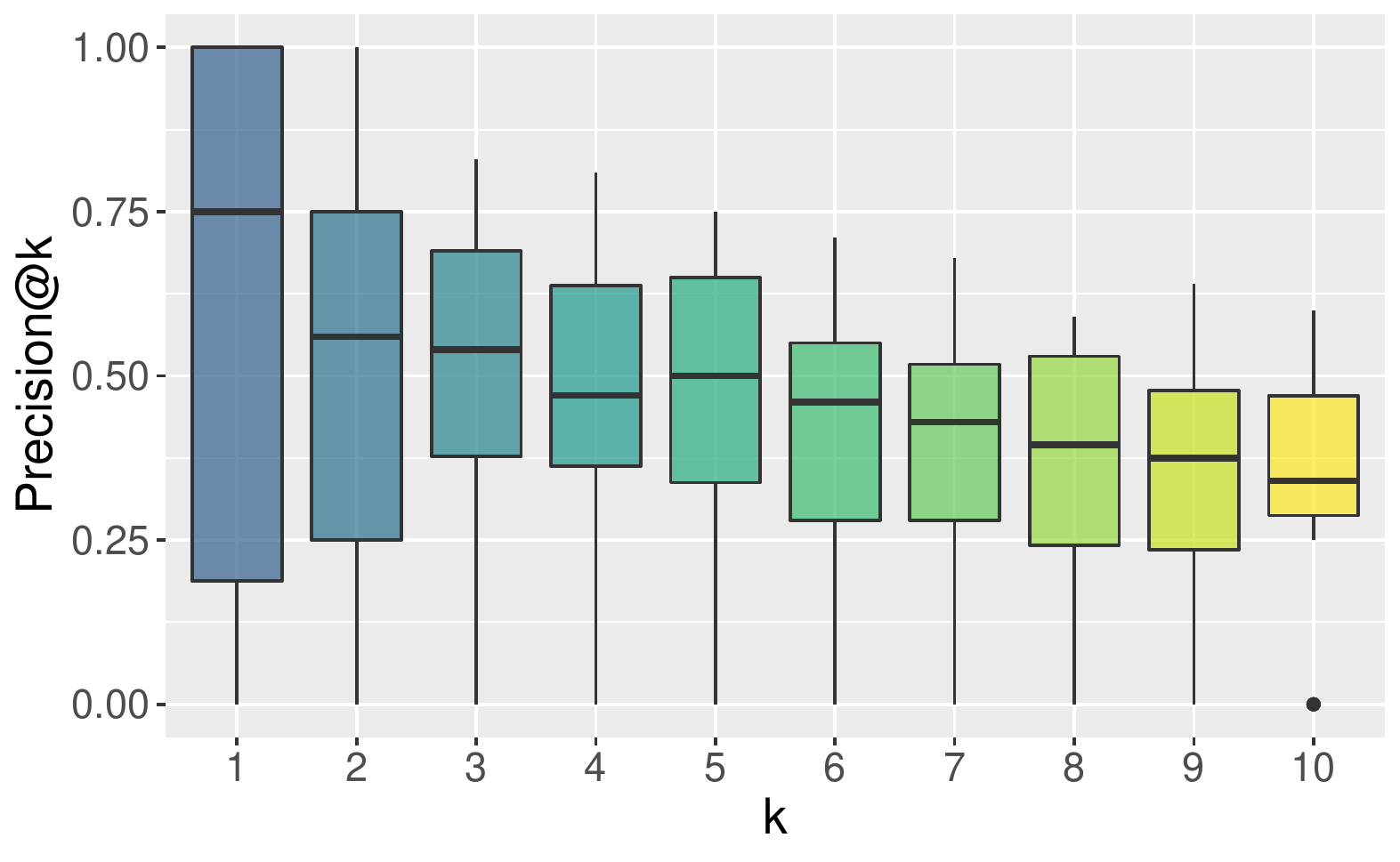} }}%
    \subfloat[English dataset]{{\includegraphics[width=0.5\textwidth]{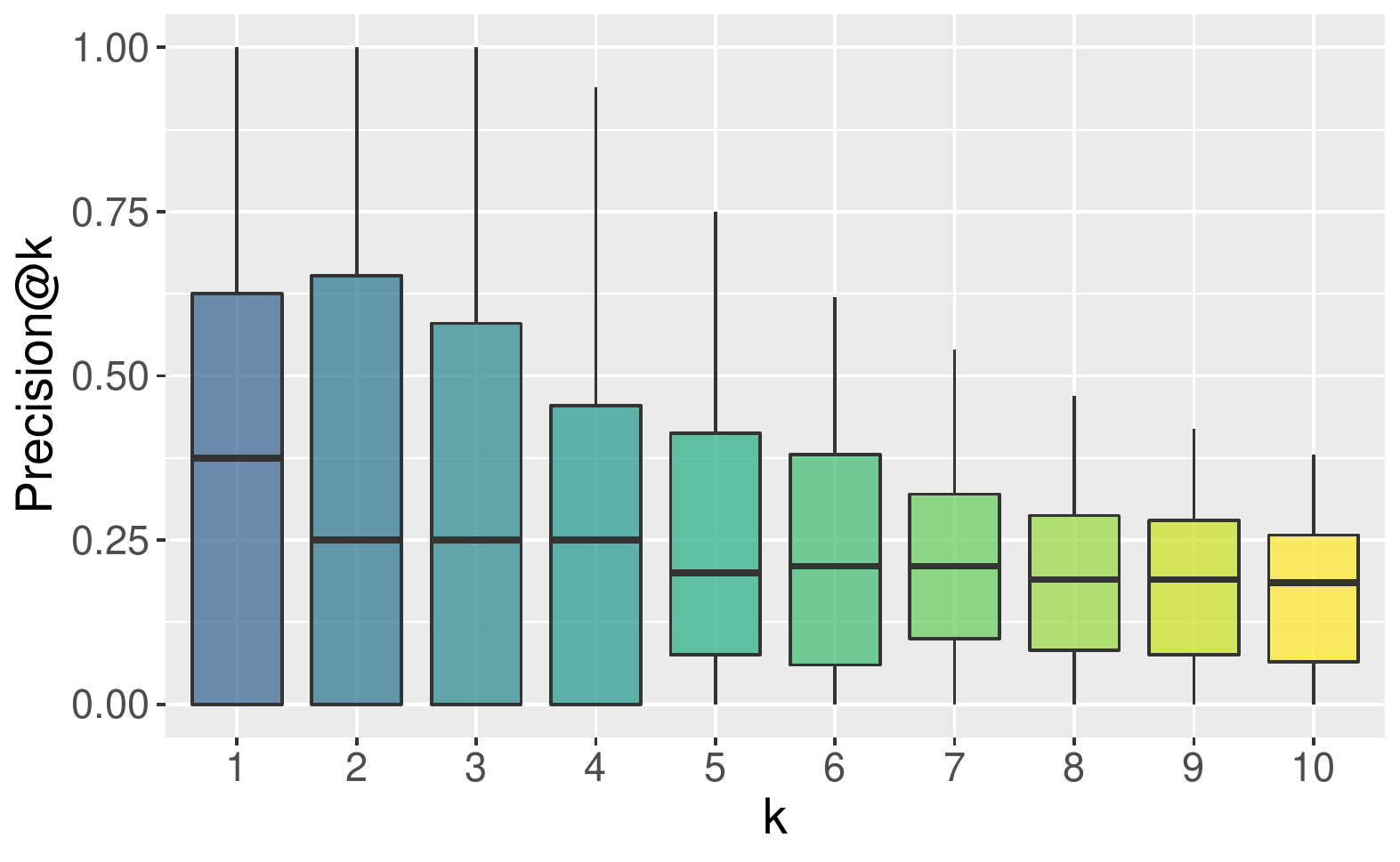} }}%
    
\end{figure*}

\begin{table*}[]
\centering
\small
\caption{Words selected for the manual evaluation.}
\label{tab:top20}
\begin{tabular}{cl}
\toprule
\textbf{Language} & \textbf{Word List} \\
\midrule
English & \begin{tabular}[c]{@{}l@{}}falanga, venerability, ambulatory, emeritus, salutatory address, eigenvalue of a matrix, \\ liposuction, moppet, dinette, snoek, to fancify, to google, to expense, to porcelainize, \\ to junketeer,  to delist, to podcast, to deglaze, to shoetree, to headquarter\end{tabular} \\
\midrule
Russian & \begin{tabular}[c]{@{}l@{}}\cyrins{барабашка}, \cyrins{листинг}, \cyrins{стихосложение}, \cyrins{аукционист}, \cyrins{точилка},  \cyrins{гиперреализм},\\ \cyrins{серология},  \cyrins{огрызок},  \cyrins{фен},  \cyrins{марикультура},  \cyrins{уломать},  \cyrins{отфотошопить}, \cyrins{тяпнуть}, \\ \cyrins{растушевать},  \cyrins{завраться},  \cyrins{леветь},  \cyrins{мозолить},  \cyrins{загоститься},  \cyrins{распеваться},    \cyrins{оплавить} 
\end{tabular}       \\
\bottomrule
\end{tabular}

\end{table*}

In order to check how useful the predicted synsets are for a human annotator (i.e. if a short list of possible hypernyms can speed up the manual extension of a taxonomy), we conduct the manual evaluation of 10 random nouns and 10 random verbs for both languages (the words are listed in Table \ref{tab:top20}). We focus on worse-quality cases and thus select words whose MAP score is below 1. Annotators with the expertise in the field and the knowledge of English and Russian were provided with guidelines and asked to evaluate the outputs from our best-performing system. Each word was labelled by 4 expert annotators, Fleiss's kappa is $0.63$ (substantial agreement) for both datasets.

We compute Precision@k score (the share of correct answers in the generated lists from position 1 to $k$) for $k$ from 1 to 10, shown in Figure \ref{fig:precision}. We can see that even for words with MAP below 1 our model manages to extract useful hypernyms.

\section{Conclusions}
\label{sec:conclusions}

In this work, we performed a large-scale computational study of various methods for taxonomy enrichment. 
We also presented datasets for studying diachronic evolution of wordnets for English and Russian, extending the monolingual setup of the RUSSE’2020 shared task \cite{nikishina2020taxonomy}
with a larger Russian dataset and similar English versions.

We presented a new taxonomy enrichment method called DWRank, which combines distributional information and the information extracted from Wiktionary outperforming the baseline method from \cite{nikishina-etal-2020-studying} on the English datasets. We also presented its extensions: DWRank-Graph and DWRank-Meta which use graph and meta- embeddings via a common interface. 

We also  explored   the   benefits   of   meta-embeddings (combinations of embeddings) and graph embeddings for the task of taxonomy enrichment.
On the Russian datasets DWRank-Meta performed best using fastText and word2vec word embeddings. For the English dataset the combination of word (fastText, word2vec and GloVe) and graph (TADW) embeddings demonstrated the best performance.

In this paper we also presented WBSR --- a method for taxonomy extension which leverages the information from the Web. This approach was the current state of the art in the task for the Russian language. Now it lags significantly behind the new DWRank-meta approaches.



According to our experiments, word vector representations are simple, powerful, and extremely effective instrument for taxonomy enrichment, as the contexts (in a broad sense) extracted from the pre-trained word embeddings (fastText, word2vec, GloVe) and their combination are sufficient to attach new words to the taxonomy. TADW embeddings are also useful and efficient for the taxonomy enrichment task and in combination with the fastText, word2vec and GloVe approaches demonstrate SOTA results for the English language and compatible results for the Russian language.

Error analysis also reveals that the correct synsets identified by graph-based models are usually retrieved by the fastText-based model alone.
This makes graphs representations mostly irrelevant and excessive. Nonetheless, there exist cases where graph representations  were able to identify correctly some hypernyms which were not captured by fastText.

Despite the mixed results of the application of graph-based methods, we propose further exploration of the graph-based features as the existing resource contains principally different and complementary information to the distributional signal contained in text corpora. One way to improve their performance, may be to use more sophisticated  non-linear projection transformations from word to graph embeddings. Another promising way in our opinion is to explore other types of meta-embeddings to mix word and graph signals, e.g.  GraphGlove \cite{ryabinin2020embedding}. Moreover, we find it promising to experiment with temporal embeddings such of those of \cite{GoelKazemiBrubakerPoupart2020} for the taxonomy enrichment task. 

Last but not least, we plan to explore methods that do not rely on the set of pre-defined candidates for inclusion in a taxonomy, as generation and mining of such a set may be a challenging problem in its own. 


\section*{Acknowledgments}
The participation of Mikhail Tikhomirov in the reported study (experiments with meta-embeddings) was funded by RFBR, project number 19-37-90119.
The work of Natalia Loukachevitch  (preparation of RuWordNet data for the experiments, mappings of the RuWordNet synsets to English WordNet Linked Open
Data (LOD) Inter-Lingual Index (ILI))  is supported by the Russian Science Foundation (project 20-11-20166). The work of Alexander Panchenko, Varvara Logacheva, and Irina Nikishina was conducted in the framework of joint MTS-Skoltech laboratory.

\nocite{*} 
\bibliographystyle{ios1}           
\bibliography{bibliography}        

%

\clearpage
\onecolumn
\phantomsection
\appendix
\section{Ranked Examples and All Results}\label{appendix}

In this Appendix we provide Tables \ref{tab:examples_en}, \ref{tab:examples_en2} and \ref{tab:examples_ru} with the examples that demonstrate the input and the output formats of the models as well as Tables \ref{tab:all_en} and \ref{tab:all_rus} that show the performance of all models for both English and Russian datasets. 

From both tables \ref{tab:examples_en} and \ref{tab:examples_en2} (underlined bold text denotes predictions of the model from the ground truth) we can see that at least one of the correct candidates usually appears in the list of candidates from word embeddings (first part of the table), whereas among candidates from graph embeddings we do not see any decent synsets. Poincaré embeddings retrieved by aggregating words from fastText provide too broad concepts which are clearly too far from the correct answers (``activity.n.01'', ``exposure.n.03'', ``action.n.02''). Node2vec embeddings are both semantically far and abstract. GraphSAGE is sticking to the word ``play'' and are too far from the correct answers in general. TADW manages to predict the correct synset ``therapy.n.01'' in the list of candidates, however, its position is much lower than the positions of the same synsets among the candidates provided by word embeddings-based systems.

The candidates for the words ``play therapy'' and ``eyewitness'' provided by models based on text embeddings do contain at least one true answer in the list. The position of such words varies from the forth to the sixth position. 

DWRank-Meta on both word and graph embeddings may provide the results that improve the ranking, e.g. ``therapy.n.01'' is the correct candidate on the first place in the list for both AAEME triple loss (fastText, word2vec and GloVe) approach and for the AAEME (fastText, word2vec, Glove, TASDW) approach. For the word ``eyewitness'' the DWRank-Meta models on words and graphs the true candidates are placed in the worse positions, however, they still win before the DWRank-GRaph approach.

\begin{table*}[ht!]
\centering
\small
\caption{Examples of predictions for noun from the English v 1.6-3.0 dataset with various models.}
\label{tab:examples_en}
\resizebox{0.8\textwidth}{!}{
\begin{tabular}{cccc}
\toprule
\multicolumn{4}{c}{\bf play therapy}  \\

\multicolumn{4}{c}{psychotherapy.n.02, therapy.n.01}
\\\midrule
 \cellcolor[HTML]{EFEFEF} fastText (baseline)           & fastText (DWRank)           & \begin{tabular}[c]{@{}c@{}}    AAEME triplet loss \\ (fastText + word2vec + \\ GloVe) \end{tabular} \cellcolor[HTML]{EFEFEF}   &  \begin{tabular}[c]{@{}c@{}} AAEME \\(fastText + word2vec + \\ GloVe + TADW)  \end{tabular}      \\            \midrule
 \cellcolor[HTML]{EFEFEF} play.n.03 & play.n.03                    &  \cellcolor[HTML]{EFEFEF}\uline{\bf therapy.n.01 }  & \uline{\bf therapy.n.01   }     \\
 \cellcolor[HTML]{EFEFEF} play.n.01 & activity.n.01             &  \cellcolor[HTML]{EFEFEF}activity.n.01                 &  medical\_care.n.01   \\
 \cellcolor[HTML]{EFEFEF} baseball\_play.n.01 & plan\_of\_action.n.01                     & \cellcolor[HTML]{EFEFEF}medical\_care.n.01  & play.n.03         \\
 \cellcolor[HTML]{EFEFEF} \uline{\bf therapy.n.01} & \uline{\bf therapy.n.01}                & \cellcolor[HTML]{EFEFEF}diversion.n.01         & play.n.01      \\
activity.n.01 \cellcolor[HTML]{EFEFEF} & play.n.01             & \cellcolor[HTML]{EFEFEF} play.n.01  &  activity.n.01    \\
 \cellcolor[HTML]{EFEFEF} diversion.n.01 & dramatic\_composition.n.01               & \cellcolor[HTML]{EFEFEF}act.n.02     & dramatic\_composition.n.01            \\
 \cellcolor[HTML]{EFEFEF} plan\_of\_action.n.01 & outdoor\_game.n.01        & \cellcolor[HTML]{EFEFEF}\uline{\bf psychotherapy.n.02}  & plan\_of\_action.n.01     \\
 \cellcolor[HTML]{EFEFEF} action.n.01 & medical\_care.n.01            & \cellcolor[HTML]{EFEFEF}play.n.03 & show.n.04  \\
 \cellcolor[HTML]{EFEFEF} action.n.02 & golf.n.01                         & \cellcolor[HTML]{EFEFEF}dramatic\_composition.n.01         &  treatment.n.01 \\
 \cellcolor[HTML]{EFEFEF} dramatic\_composition.n.01 & diversion.n.01          & \cellcolor[HTML]{EFEFEF}behaviour\_therapy.n.01 &     act.n.02 \\ \midrule
\midrule
 GraphSAGE & \cellcolor[HTML]{EFEFEF}  TADW              & Poincaré   &  node2vec \cellcolor[HTML]{EFEFEF}                                         \\ \midrule
 baseball\_play.n.01  & \cellcolor[HTML]{EFEFEF}   play.n.03                                                   & activity.n.01     & \cellcolor[HTML]{EFEFEF} presentation.n.03                                                  \\
 play.n.03   & \cellcolor[HTML]{EFEFEF}play.n.01                                                               &   exposure.n.03       & \cellcolor[HTML]{EFEFEF} presentation.n.01                                                  \\
 play.n.01  & \cellcolor[HTML]{EFEFEF}plan\_of\_action.n.01                                                         &    rejection.n.01        & \cellcolor[HTML]{EFEFEF} operation.n.01                                              \\
 squeeze\_play.n.02   & \cellcolor[HTML]{EFEFEF}        dramatic\_composition.n.01      &         agreement.n.06                                          & \cellcolor[HTML]{EFEFEF} performance.n.02                         \\
 activity.n.01   & \cellcolor[HTML]{EFEFEF}activity.n.01                                    & light\_unit.n.01                                             & \cellcolor[HTML]{EFEFEF} contact.n.01           \\
 diversion.n.01  & \cellcolor[HTML]{EFEFEF}       baseball\_play.n.01                                                         & blessing.n.01                                            & \cellcolor[HTML]{EFEFEF} exposure.n.03        \\
 dramatic\_composition.n.01   & \cellcolor[HTML]{EFEFEF}  show.n.04                                            & vulnerability.n.02                                            & \cellcolor[HTML]{EFEFEF} activity.n.01         \\
  attempt.n.01  & \cellcolor[HTML]{EFEFEF}diversion.n.01                                                       &          action.n.02                   & \cellcolor[HTML]{EFEFEF} exposure.n.08                               \\
  plan\_of\_action.n.01  & \cellcolor[HTML]{EFEFEF}use.n.01                                                            &       influence.n.02                                                 & \cellcolor[HTML]{EFEFEF}  union.n.04    \\
 play.n.17   & \cellcolor[HTML]{EFEFEF}action.n.02                                                               &  assent.n.01    & \cellcolor[HTML]{EFEFEF} exposure.n.06                                                      \\ \bottomrule                    
\end{tabular}}
\end{table*}

\begin{table*}[ht!]
\centering
\small
\caption{Examples of predictions for nouns from the English v 1.6-3.0 dataset with various models.}
\label{tab:examples_en2}
\resizebox{0.8\textwidth}{!}{
\begin{tabular}{cccc}
\toprule 
 \multicolumn{4}{c}{\bf Ramadan}                                                         \\     \multicolumn{4}{c}{islamic\_calendar\_month.n.01, calendar\_month.n.01, fast.n.01, abstinence.n.02}                                                                                                                                                                                                 \\ \midrule
             \cellcolor[HTML]{EFEFEF} fastText (baseline)           & fastText (DWRank)           &
              \begin{tabular}[c]{@{}c@{}}    AAEME triplet loss \\ (fastText + word2vec + \\ GloVe) \end{tabular} \cellcolor[HTML]{EFEFEF}   &  \begin{tabular}[c]{@{}c@{}} AAEME \\(fastText + word2vec + \\ GloVe + TADW)  \end{tabular}                                         \\\midrule
 \cellcolor[HTML]{EFEFEF}       \uline{\bf islamic\_calendar\_month.n.01}    &    \uline{\bf islamic\_calendar\_month.n.01}           & \cellcolor[HTML]{EFEFEF}calendar.n.01                &   \uline{\bf islamic\_calendar\_month.n.01} \\
 \cellcolor[HTML]{EFEFEF}\uline{\bf calendar\_month.n.01  }     & \uline{\bf calendar\_month.n.01  }    &  \cellcolor[HTML]{EFEFEF}\uline{\bf islamic\_calendar\_month.n.01} &   \uline{\bf calendar\_month.n.01  }             \\

 \cellcolor[HTML]{EFEFEF}holiday.n.02
               &       time\_period.n.01       & \cellcolor[HTML]{EFEFEF} \uline{\bf calendar\_month.n.01  }                                                      &   time\_period.n.01       \\

 \cellcolor[HTML]{EFEFEF}religious\_holiday.n.01         & calendar.n.01  & \cellcolor[HTML]{EFEFEF}         lunar\_calendar.n.01                                               &   calendar.n.01          \\
 \cellcolor[HTML]{EFEFEF}\uline{\bf abstinence.n.02}   &        islam.n.01    & \cellcolor[HTML]{EFEFEF} time\_period.n.01                                  &  muslim.n.01\\\
 \cellcolor[HTML]{EFEFEF}  hindu\_calendar\_month.n.01      &    lunar\_calendar.n.01     & \cellcolor[HTML]{EFEFEF}religionist.n.01  &   religionist.n.01       \\
 \cellcolor[HTML]{EFEFEF} day.n.04          &   asian.n.01             & \cellcolor[HTML]{EFEFEF}muslim.n.01                                                     &   religious\_holiday.n.01       \\
 \cellcolor[HTML]{EFEFEF} sacred\_text.n.01           &          religion.n.02    & \cellcolor[HTML]{EFEFEF} religion.n.02                                                        &      holiday.n.02     \\
 \cellcolor[HTML]{EFEFEF}       god.n.01     &  muslim.n.01          &  \cellcolor[HTML]{EFEFEF}islam.n.01                              &  lunar\_calendar.n.01      \\
 \cellcolor[HTML]{EFEFEF}   place\_of\_worship.n.01   &            holiday.n.02  & \cellcolor[HTML]{EFEFEF}person.n.01                                                      & islam.n.01    \\ \midrule\midrule

 graphSAGE & \cellcolor[HTML]{EFEFEF}TADW                                                      &    
              Poincaré & \cellcolor[HTML]{EFEFEF} node2vec \\\midrule
    \uline{\bf islamic\_calendar\_month.n.01} & \cellcolor[HTML]{EFEFEF}\uline{\bf islamic\_calendar\_month.n.01}                                                 & time\_period.n.01     &                            \cellcolor[HTML]{EFEFEF} \uline{\bf calendar\_month.n.01}     \\
   \uline{\bf calendar\_month.n.01} & \cellcolor[HTML]{EFEFEF}\uline{\bf calendar\_month.n.01}      &  \uline{\bf islamic\_calendar\_month.n.01}   &  \cellcolor[HTML]{EFEFEF}\uline{\bf islamic\_calendar\_month.n.01} \\

    arab.n.01 &  \cellcolor[HTML]{EFEFEF}time.n.02                        & religion.n.02 &  \cellcolor[HTML]{EFEFEF} revolutionary\_calendar\_month.n.01 \\muslim.n.01
   & \cellcolor[HTML]{EFEFEF}religious\_holiday.n.01                                        & time.n.02   &    \cellcolor[HTML]{EFEFEF} islam.n.01  \\semite.n.01
   & \cellcolor[HTML]{EFEFEF}calendar.n.01                         &    measure.n.03          &     \cellcolor[HTML]{EFEFEF}  time\_period.n.01\\religionist.n.01
   & \cellcolor[HTML]{EFEFEF}time\_period.n.01 & \uline{\bf calendar\_month.n.01}  &    \cellcolor[HTML]{EFEFEF} hindu\_calendar\_month.n.01\\god.n.01
   & \cellcolor[HTML]{EFEFEF}  holiday.n.02                                 & term.n.02  &        \cellcolor[HTML]{EFEFEF}    muharram.n.01\\saint.n.02
    & \cellcolor[HTML]{EFEFEF}   lunar\_calendar.n.01                                       & year.n.01   &  \cellcolor[HTML]{EFEFEF}shawwal.n.01\\zoysia.n.01
  & \cellcolor[HTML]{EFEFEF}god.n.01                                                      &  time\_off.n.01  &      \cellcolor[HTML]{EFEFEF}rabi\_i.n.01   \\islam.n.01
     & \cellcolor[HTML]{EFEFEF} jewish\_holy\_day.n.01                                                             &   leisure.n.01 &      \cellcolor[HTML]{EFEFEF}rajab.n.01   \\  \bottomrule                                   
\end{tabular}}
\vspace{0.5cm}
\small
\caption{Examples of predictions for verbs from the English v 1.6-3.0 dataset with various models.}
\label{tab:examples_ru}
\resizebox{0.7\textwidth}{!}{
\begin{tabular}{cccc}
\toprule 
 \multicolumn{4}{c}{\bf eyewitness}                                                         \\     \multicolumn{4}{c}{witness.v.01, watch.v.01}                                                                                                                                                                                                 \\ \midrule
             \cellcolor[HTML]{EFEFEF} fastText (baseline)           & fastText (DWRank)           &
              \begin{tabular}[c]{@{}c@{}}    AAEME triplet loss \\ (fastText + word2vec + \\ GloVe) \end{tabular} \cellcolor[HTML]{EFEFEF}   &  \begin{tabular}[c]{@{}c@{}} AAEME \\(fastText + word2vec + \\ GloVe + TADW)  \end{tabular}                                         \\\midrule
 \cellcolor[HTML]{EFEFEF}        be.v.01    &     inform.v.01           & \cellcolor[HTML]{EFEFEF}inform.v.01                & inform.v.01   \\
 \cellcolor[HTML]{EFEFEF} be.v.03 & testify.v.02 &  \cellcolor[HTML]{EFEFEF}testify.v.02 & testify.v.02            \\

 \cellcolor[HTML]{EFEFEF}
    be.v.08           & communicate.v.02             & \cellcolor[HTML]{EFEFEF} communicate.v.02                                                        & confirm.v.02          \\

 \cellcolor[HTML]{EFEFEF} \uline{\textbf{watch.v.01}}        & announce.v.01 & \cellcolor[HTML]{EFEFEF} see.v.10                                                       & communicate.v.02            \\
 \cellcolor[HTML]{EFEFEF}  testify.v.01 & confirm.v.02           & \cellcolor[HTML]{EFEFEF}confirm.v.02                                             & \uline{\bf watch.v.01}  \\
 \cellcolor[HTML]{EFEFEF}     testify.v.02   & \uline{\bf watch.v.01}         & \cellcolor[HTML]{EFEFEF} \uline{\bf watch.v.01}  & be.v.01          \\
 \cellcolor[HTML]{EFEFEF}  man.v.02           & testify.v.01               & \cellcolor[HTML]{EFEFEF}    witness.v.02                                                  & affirm.v.03         \\
 \cellcolor[HTML]{EFEFEF}    talk.v.01        & affirm.v.03              & \cellcolor[HTML]{EFEFEF}affirm.v.03                                                           & testify.v.01           \\
 \cellcolor[HTML]{EFEFEF}       guard.v.01     &  report.v.03           &  \cellcolor[HTML]{EFEFEF} testify.v.01                             & reject.v.01       \\
 \cellcolor[HTML]{EFEFEF}   confirm.v.01   &        record.v.01      & \cellcolor[HTML]{EFEFEF}  verify.v.01                                                    & experience.v.01    \\ \midrule\midrule

 graphSAGE & \cellcolor[HTML]{EFEFEF}TADW                                                      &    
              Poincaré & \cellcolor[HTML]{EFEFEF} node2vec \\\midrule
   see.v.05  & \cellcolor[HTML]{EFEFEF}                              inform.v.01                    &   confirm.v.02     &      affirm.v.03                       \cellcolor[HTML]{EFEFEF}      \\
 testify.v.02  & \cellcolor[HTML]{EFEFEF}testify.v.02       & examine.v.02    & confirm.v.02 \cellcolor[HTML]{EFEFEF}\\

  err.v.01   & communicate.v.02 \cellcolor[HTML]{EFEFEF}                                                               & affirm.v.03 & understand.v.02   \cellcolor[HTML]{EFEFEF} \\
 confirm.v.01  & \cellcolor[HTML]{EFEFEF}declare.v.01                                                               &  testify.v.02  & uphold.v.03   \cellcolor[HTML]{EFEFEF}  \\
 pronounce.v.02  & \cellcolor[HTML]{EFEFEF}announce.v.01                                    &             inform.v.01 &    determine.v.08 \cellcolor[HTML]{EFEFEF}  \\
  idealize.v.01  & \cellcolor[HTML]{EFEFEF}testify.v.01 &  justify.v.02 & stay\_in\_place.v.01     \cellcolor[HTML]{EFEFEF} \\
  judge.v.02   & \cellcolor[HTML]{EFEFEF}record.v.01                                                           &  declare.v.01 &    justify.v.02    \cellcolor[HTML]{EFEFEF}    \\
 negate.v.03   & \cellcolor[HTML]{EFEFEF}                                       \uline{\bf watch.v.01}                      &   validate.v.03 & fall\_asleep.v.01   \cellcolor[HTML]{EFEFEF}\\
reason.v.01  & \cellcolor[HTML]{EFEFEF}    report.v.03                                                    &                            uphold.v.03 &      resettle.v.01 \cellcolor[HTML]{EFEFEF}   \\
   disbelieve.v.01  & \cellcolor[HTML]{EFEFEF}  report.v.01                                                             &  testify.v.01 &   settle.v.04    \cellcolor[HTML]{EFEFEF}   \\  \bottomrule                                   
\end{tabular}}
\end{table*}

\begin{table*}[ht!]
\centering
\small
\caption{Examples of predictions for verbs from the English v 1.6-3.0 dataset with various models.}
\label{tab:examples_en3}
\resizebox{0.65\textwidth}{!}{
\begin{tabular}{cccc}
\toprule 
 \multicolumn{4}{c}{\bf theologise}                                                       \\     \multicolumn{4}{c}{cover.v.05, broach.v.01, chew\_over.v.01, think.v.03}                                                                                                                                                                                                 \\ \midrule
             \cellcolor[HTML]{EFEFEF} fastText (baseline)           & fastText (DWRank)           &
              \begin{tabular}[c]{@{}c@{}}    AAEME triplet loss \\ (fastText + word2vec + \\ GloVe) \end{tabular} \cellcolor[HTML]{EFEFEF}   &  \begin{tabular}[c]{@{}c@{}} AAEME \\(fastText + word2vec + \\ GloVe + TADW)  \end{tabular}                                         \\\midrule
 \cellcolor[HTML]{EFEFEF}  change.v.01         &  change.v.01              & \cellcolor[HTML]{EFEFEF}  change.v.01             &  change.v.01 \\
 \cellcolor[HTML]{EFEFEF}match.v.01  & preface.v.01 &  \cellcolor[HTML]{EFEFEF}preface.v.01 & preface.v.01             \\

 \cellcolor[HTML]{EFEFEF}
           date.v.03 &       make.v.03     & \cellcolor[HTML]{EFEFEF}convert.v.05                                                        &     equal.v.03      \\

 \cellcolor[HTML]{EFEFEF}   reconstruct.v.01      & date.v.03 & \cellcolor[HTML]{EFEFEF}  match.v.01                                                    &   date.v.03         \\
 \cellcolor[HTML]{EFEFEF}determine.v.03   &      equal.v.03      & \cellcolor[HTML]{EFEFEF}    chronologize.v.01                                        &  reconstruct.v.01 \\
 \cellcolor[HTML]{EFEFEF}  make.v.03      &    reason.v.01    & \cellcolor[HTML]{EFEFEF}reason.v.01 &         reason.v.01 \\
 \cellcolor[HTML]{EFEFEF}   speculate.v.01         &     chronologize.v.01          & \cellcolor[HTML]{EFEFEF}   change\_state.v.01                                                 &   convert.v.05      \\
 \cellcolor[HTML]{EFEFEF}  convert.v.05         &      match.v.01         & \cellcolor[HTML]{EFEFEF} represent.v.09                                                          &  formulate.v.03         \\
 \cellcolor[HTML]{EFEFEF} reason.v.01           &   film.v.02         &  \cellcolor[HTML]{EFEFEF}make.v.03                              &      reflect.v.04  \\
 \cellcolor[HTML]{EFEFEF} automatize.v.01     &      formulate.v.03       & \cellcolor[HTML]{EFEFEF}change.v.02                                                      & film.v.02    \\ \midrule\midrule

 graphSAGE & \cellcolor[HTML]{EFEFEF}TADW                                                      &    
              Poincaré & \cellcolor[HTML]{EFEFEF} node2vec \\\midrule
   process.v.02  & \cellcolor[HTML]{EFEFEF}           change.v.01                                      &    state.v.01    &              settle.v.04               \cellcolor[HTML]{EFEFEF}      \\
affect.v.01   & \cellcolor[HTML]{EFEFEF} preface.v.01      &  speculate.v.01   &  \cellcolor[HTML]{EFEFEF}stay\_in\_place.v.01\\

  change.v.01   &  \cellcolor[HTML]{EFEFEF}                                                     date.v.03          & reason.v.01 &    \cellcolor[HTML]{EFEFEF}understand.v.02 \\
  dive.v.01 & \cellcolor[HTML]{EFEFEF}                                                         film.v.02      &  preface.v.01  &  \cellcolor[HTML]{EFEFEF}study.v.03  \\
   tame.v.01 & \cellcolor[HTML]{EFEFEF}reconstruct.v.01                                    &    match.v.01          &     \cellcolor[HTML]{EFEFEF}resettle.v.01  \\
 sensitize.v.02  & \cellcolor[HTML]{EFEFEF}equal.v.03 & generalize.v.01  &      \cellcolor[HTML]{EFEFEF} fall\_asleep.v.01\\
   convert.v.01  & \cellcolor[HTML]{EFEFEF}                                                  convert.v.05         &  announce.v.02 &        \cellcolor[HTML]{EFEFEF}speculate.v.01    \\
  estimate.v.01 & \cellcolor[HTML]{EFEFEF}                                                formulate.v.03  &  express.v.02  &  \cellcolor[HTML]{EFEFEF}discover.v.07\\
subject.v.01  & \cellcolor[HTML]{EFEFEF}                                                   reason.v.01  &     add.v.02                       &       \cellcolor[HTML]{EFEFEF}explicate.v.02   \\
 compound.v.05   & \cellcolor[HTML]{EFEFEF}                                                        commemorate.v.03       & equal.v.01 &       \cellcolor[HTML]{EFEFEF}  behave.v.02 \\  \bottomrule                                   
\end{tabular}}
\vspace{0.5cm}
\small
\caption{Examples of predictions for verbs from the English v 1.6-3.0 dataset with various models.}
\label{tab:examples_ru2}
\resizebox{0.65\textwidth}{!}{
\begin{tabular}{cccc}
\toprule 
 \multicolumn{4}{c}{\bf immunise}                                                         \\     \multicolumn{4}{c}{protect.v.01, defend.v.02, inject.v.01, administer.v.04}                                                                                                                                                                                                 \\ \midrule
             \cellcolor[HTML]{EFEFEF} fastText (baseline)           & fastText (DWRank)           &
              \begin{tabular}[c]{@{}c@{}}    AAEME triplet loss \\ (fastText + word2vec + \\ GloVe) \end{tabular} \cellcolor[HTML]{EFEFEF}   &  \begin{tabular}[c]{@{}c@{}} AAEME \\(fastText + word2vec + \\ GloVe + TADW)  \end{tabular}                                         \\\midrule
 \cellcolor[HTML]{EFEFEF} indoctrinate.v.01          & indoctrinate.v.01               & \cellcolor[HTML]{EFEFEF} teach.v.01              & \uline{\bf inject.v.01}  \\
 \cellcolor[HTML]{EFEFEF}\uline{\bf inject.v.01}  & \uline{\bf  protect.v.01}  &  \cellcolor[HTML]{EFEFEF}treat.v.01 &  remove.v.01           \\

  \uline{\bf defend.v.02}\cellcolor[HTML]{EFEFEF}
            &    \uline{\bf defend.v.02}   & \cellcolor[HTML]{EFEFEF} insert.v.02                                                       &     inform.v.01      \\

 \cellcolor[HTML]{EFEFEF}remove.v.01         & teach.v.01 & \cellcolor[HTML]{EFEFEF}prevent.v.01                                                      &    change.v.01        \\
 \cellcolor[HTML]{EFEFEF}treat.v.01   &    prevent.v.02        & \cellcolor[HTML]{EFEFEF}    isolate.v.01                                        & better.v.02 \\
 \cellcolor[HTML]{EFEFEF} destroy.v.01       &    \uline{\bf inject.v.01}    & \cellcolor[HTML]{EFEFEF}inoculate.v.01 &   insert.v.02       \\
 \cellcolor[HTML]{EFEFEF}   prevent.v.02         &      insert.v.02         & \cellcolor[HTML]{EFEFEF} indoctrinate.v.01                                                   &    defend.v.02     \\
 \cellcolor[HTML]{EFEFEF}insert.v.02           &    defend.v.01           & \cellcolor[HTML]{EFEFEF} kill.v.01                                                          &    kill.v.01       \\
 \cellcolor[HTML]{EFEFEF}     teach.v.01       & kill.v.01           &  \cellcolor[HTML]{EFEFEF}        change.v.01                      &      indoctrinate.v.01  \\
 \cellcolor[HTML]{EFEFEF}\uline{\bf administer.v.04}     &  discriminate.v.02           & \cellcolor[HTML]{EFEFEF} enable.v.01                                                     & protect.v.01    \\ \midrule\midrule

 graphSAGE & \cellcolor[HTML]{EFEFEF}TADW                                                      &    
              Poincaré & \cellcolor[HTML]{EFEFEF} node2vec \\\midrule
  \uline{\bf defend.v.02} & \cellcolor[HTML]{EFEFEF}\uline{\bf protect.v.01}                                               &  teach.v.01      &                             \cellcolor[HTML]{EFEFEF}receive.v.01      \\
  \uline{\bf protect.v.01} & \cellcolor[HTML]{EFEFEF}  indoctrinate.v.01     &  insert.v.02   &  \cellcolor[HTML]{EFEFEF}insert.v.02\\

  act.v.01   &  \cellcolor[HTML]{EFEFEF}teach.v.01                                                               & inform.v.01 &    \cellcolor[HTML]{EFEFEF}stay\_in\_place.v.01 \\
   negociate.v.01 & \cellcolor[HTML]{EFEFEF} \uline{\bf defend.v.02}                                                              &    treat.v.01 &  \cellcolor[HTML]{EFEFEF} get.v.01 \\
  attack.v.03 & \cellcolor[HTML]{EFEFEF} prevent.v.02                                   &        train.v.01      &     \cellcolor[HTML]{EFEFEF} \uline{\bf inject.v.01} \\
  prevent.v.02 & \cellcolor[HTML]{EFEFEF} \uline{\bf inject.v.01} &  deceive.v.02 &      \cellcolor[HTML]{EFEFEF}fall\_asleep.v.01 \\
   insert.v.02  & \cellcolor[HTML]{EFEFEF}prevent.v.01                                                           &  indoctrinate.v.01 &        \cellcolor[HTML]{EFEFEF} accept.v.02    \\
 demilitarize.v.01  & \cellcolor[HTML]{EFEFEF} insert.v.02                                                 &  misinform.v.01  &  \cellcolor[HTML]{EFEFEF}settle.v.04 \\
  disarm.v.02 & \cellcolor[HTML]{EFEFEF}isolate.v.01                                                     &     gull.v.02                       &       \cellcolor[HTML]{EFEFEF} resettle.v.01  \\
   foreswear.v.02 & \cellcolor[HTML]{EFEFEF} discriminate.v.02                                                              & interact.v.01 &       \cellcolor[HTML]{EFEFEF}  discover.v.07 \\  \bottomrule                                   
\end{tabular}}
\end{table*}

\begin{table*}[ht]
\centering
\small
\caption{MAP scores for the taxonomy enrichment methods for the English datasets. Numbers \textbf{in bold} show the best model within the category, \underline{\smash{\bf underlined}} numbers denote the best score across all the models. The combination of word embeddings (fastText, word2vec, GloVe) is denoted as \textit{words}.}

\label{tab:all_en}
\resizebox{\textwidth}{!}{
\begin{tabular}{l|rrr|rrr}
\toprule
\multicolumn{1}{c|}{\multirow{2}{*}{method}}                                     & \multicolumn{3}{c|}{Nouns}                                                               & \multicolumn{3}{c}{Verbs}                                                               \\
\multicolumn{1}{c|}{}                                                            & \multicolumn{1}{c}{1.6-3.0} & \multicolumn{1}{c}{1.7-3.0} & \multicolumn{1}{c|}{2.0-3.0} & \multicolumn{1}{c}{1.6-3.0} & \multicolumn{1}{c}{1.7-3.0} & \multicolumn{1}{c}{2.0-3.0} \\
\midrule
\multicolumn{7}{c}{\bf Baseline \cite{nikishina-etal-2020-studying}}                                  \\
\midrule
fastText \cite{bojanowski-etal-2017-enriching}
& \textbf{0.338$\pm$0.002}             & \textbf{0.371$\pm$0.002}             & \textbf{0.400$\pm$0.004}             & \textbf{0.270$\pm$0.007}            & \textbf{0.203$\pm$0.010}            & \textbf{0.236$\pm$0.011}                     \\
word2vec    \cite{word2vec}                                                            &         0.142$\pm$0.001                   &    0.178$\pm$0.002       &     0.164$\pm$0.004     &   0.229$\pm$0.006   & 0.155$\pm$0.008   &    0.212$\pm$0.009       \\
GloVe     \cite{pennington-etal-2014-glove}                                                           &          0.232$\pm$0.002                   &    0.188$\pm$0.001       &     0.233$\pm$0.004      &   0.146$\pm$0.005   & 0.149$\pm$0.008   &    0.191$\pm$0.010      \\ \midrule
    \multicolumn{7}{c}{\bf DWRank-Word}                                  \\
\midrule
fastText \cite{bojanowski-etal-2017-enriching} &\textbf{0.314$\pm$0.001}&\textbf{0.373$\pm$0.003}&\textbf{0.418$\pm$0.004}&\textbf{0.286$\pm$0.007}&\textbf{0.218$\pm$0.008}&\textbf{0.254$\pm$0.012}\\
word2vec \cite{word2vec} &0.244$\pm$0.001&0.271$\pm$0.003&0.298$\pm$0.004&0.099$\pm$0.005&0.118$\pm$0.008&0.141$\pm$0.010\\
GloVe \cite{pennington-etal-2014-glove} &0.283$\pm$0.001&0.329$\pm$0.003&0.377$\pm$0.004&0.182$\pm$0.007&0.159$\pm$0.008&0.203$\pm$0.011\\
\midrule
\multicolumn{7}{c}{\bf DWRank-Meta (Meta-embeddings based on Word Embeddings)}\\
\midrule
concat (\textit{words}) &\textbf{0.335$\pm$0.001}&0.386$\pm$0.003&0.386$\pm$0.003&0.270$\pm$0.007&0.194$\pm$0.009&0.226$\pm$0.011\\
SVD (\textit{words})  &0.333$\pm$0.001&\textbf{0.399$\pm$0.003}&\textbf{0.456$\pm$0.004}&0.277$\pm$0.007&0.209$\pm$0.010&0.264$\pm$0.012\\
CAEME\textit{words})  &0.321$\pm$0.001&0.386$\pm$0.003&0.448$\pm$0.005&0.278$\pm$0.007&0.205$\pm$0.008&0.266$\pm$0.015\\
AAEME (\textit{words}) &0.322$\pm$0.001&0.384$\pm$0.003&0.453$\pm$0.004&0.271$\pm$0.007&\textbf{0.218$\pm$0.008}&\textbf{0.273$\pm$0.012}\\
\midrule
CAEME triplet loss (\textit{words})  &0.332$\pm$0.001&0.394$\pm$0.003&0.451$\pm$0.004&0.273$\pm$0.007&0.205$\pm$0.007&0.256$\pm$0.013\\
AAEME triplet loss (\textit{words})  &\textbf{0.335$\pm$0.001}&0.391$\pm$0.003&0.453$\pm$0.004&\textbf{0.280$\pm$0.008}&0.212$\pm$0.007&0.262$\pm$0.014\\
\midrule


\multicolumn{7}{c}{\bf DWRank-Graph}\\
\midrule
GCN \cite{kipf2017semi} & 0.175$\pm$0.001&0.249$\pm$0.002&0.267$\pm$0.002&0.162$\pm$0.006&0.113$\pm$0.005&0.149$\pm$0.010\\
GAT \cite{velickovic2018graph} &0.000$\pm$0.000&0.252$\pm$0.002&0.000$\pm$0.000&0.081$\pm$0.003&0.064$\pm$0.004&0.000$\pm$0.000\\
GraphSAGE \cite{hamilton2017inductive} &0.214$\pm$0.001&0.282$\pm$0.002&0.224$\pm$0.003&0.127$\pm$0.004&0.114$\pm$0.004&0.090$\pm$0.008\\
TADW \cite{yang2015network} (on fastText)& \textbf{0.350$\pm$0.001}                      & \textbf{0.392$\pm$0.002}                      &\textbf{0.435$\pm$0.004}& \textbf{0.268$\pm$0.007}                      & \textbf{0.201$\pm$0.007}                      &\textbf{0.217$\pm$0.010}                     \\
Poincaré \cite{nickel2017poincare} (top-5 fastText associates)&0.185$\pm$0.001&0.211$\pm$0.002&0.229$\pm$0.002&0.208$\pm$0.006&0.147$\pm$0.006&0.172$\pm$0.012\\
node2vec \cite{node2vec-kdd2016} (top-5 fastText associates)&0.270$\pm$0.001&0.312$\pm$0.002&0.341$\pm$0.004&0.175$\pm$0.006&0.128$\pm$0.007&0.118$\pm$0.012\\
HOPE \cite{ou2016asymmetric}  & 0.000$\pm$0.000                      & 0.000$\pm$0.000                      & 0.000$\pm$0.000                      & 0.000$\pm$0.000                      & 0.000$\pm$0.000                      & 0.000$\pm$0.000                   \\\midrule

\multicolumn{7}{c}{\bf DWRank-Meta (Meta-embeddings based on Word and Graph Embeddings) }\\
\midrule
SVD (\textit{words} + node2vec)  &0.343$\pm$0.001&0.383$\pm$0.003&0.434$\pm$0.005&0.272$\pm$0.006&0.194$\pm$0.009&0.239$\pm$0.011\\
CAEME (\textit{words} + node2vec)  &0.335$\pm$0.001&0.379$\pm$0.003&0.426$\pm$0.004&0.242$\pm$0.005&0.184$\pm$0.009&0.221$\pm$0.012\\
AAEME (\textit{words} + node2vec)  &0.350$\pm$0.001&0.394$\pm$0.003&0.446$\pm$0.004&0.252$\pm$0.007&0.184$\pm$0.008&0.208$\pm$0.012\\
\midrule
SVD (\textit{words} + TADW)  &0.355$\pm$0.001&0.414$\pm$0.003&0.472$\pm$0.004& \underline{\textbf{0.288$\pm$0.007}} &0.222$\pm$0.009&\textbf{\uline{0.280$\pm$0.013}}\\
CAEME (\textit{words} + TADW)  &0.350$\pm$0.001&0.404$\pm$0.003&0.458$\pm$0.004&0.267$\pm$0.007&0.212$\pm$0.007&0.247$\pm$0.011\\
AAEME (\textit{words} + TADW)   &\textbf{\uline{0.367$\pm$0.001}}&\textbf{\uline{0.418$\pm$0.002}}&\textbf{ \uline{0.480$\pm$0.004}}&0.283$\pm$0.007&\textbf{\uline{0.227$\pm$0.007}}&0.260$\pm$0.012\\ 
\midrule
SVD (\textit{words} + GCN) &0.323$\pm$0.001&0.385$\pm$0.003&0.443$\pm$0.004&0.260$\pm$0.005&0.209$\pm$0.009&0.249$\pm$0.011\\
CAEME (\textit{words} + GCN) &0.331$\pm$0.001&0.395$\pm$0.003&0.457$\pm$0.004&0.251$\pm$0.006&0.207$\pm$0.009&0.235$\pm$0.012\\
AAEME (\textit{words} + GCN) &0.331$\pm$0.001&0.392$\pm$0.003&0.456$\pm$0.004&0.243$\pm$0.006&0.200$\pm$0.008&0.228$\pm$0.012\\
\midrule
SVD (\textit{words} + GraphSAGE)  &0.338$\pm$0.001&0.401$\pm$0.003&0.464$\pm$0.004&0.239$\pm$0.006&0.194$\pm$0.009&0.221$\pm$0.011\\
CAEME (\textit{words} + GraphSAGE) &0.323$\pm$0.001&0.382$\pm$0.003&0.435$\pm$0.004&0.200$\pm$0.006&0.170$\pm$0.007&0.202$\pm$0.01\\
AAEME (\textit{words} + GraphSAGE) &0.343$\pm$0.001&0.406$\pm$0.003&0.468$\pm$0.004&0.238$\pm$0.007&0.178$\pm$0.008&0.209$\pm$0.011\\
\midrule
\multicolumn{7}{c}{\bf State-of-the-art Approaches}\\
\midrule
WBSR (Top-1 RUSSE'2020 for nouns)                                                              & \multicolumn{1}{r}{\bf 0.333$\pm$0.002}              & \textbf{0.393$\pm$0.003}          & \textbf{0.436$\pm$0.003}               & \textbf{0.252$\pm$0.006}  & \textbf{0.206$\pm$0.011}  & \textbf{0.252$\pm$0.013}  \\
\begin{tabular}[l]{@{}l@{}}WBSR,                   no search engine features \end{tabular}                        & \multicolumn{1}{r}{0.251$\pm$0.001}              & 0.309$\pm$0.003          & 0.344$\pm$0.004             &   0.231$\pm$0.006 & 0.180$\pm$0.008 & 0.222$\pm$0.009\\
\midrule
hypo2path rev \cite{cho-etal-2020-leveraging}               &   0.264$\pm$0.001                          &    0.283$\pm$0.003      &   0.238$\pm$0.007        &     0.173$\pm$0.005 & 0.104$\pm$0.008  &  0.118$\pm$0.009   \\
hypo2path \cite{cho-etal-2020-leveraging}              &          0.252$\pm$0.002                    &      0.261$\pm$0.002     &      0.208$\pm$0.006     &   0.162$\pm$0.005 & 0.093$\pm$0.006   &   0.067$\pm$0.008  \\
hypo2path transformer              &          0.218$\pm$0.002                    &      0.229$\pm$0.002     &      0.057$\pm$0.002     &  0.140$\pm$0.003  &  0.120$\pm$0.006  &   0.100$\pm$0.008  \\\midrule
\textit{TaxoExpan} \cite{shen2020taxoexpan}      & \textit{0.004$\pm$0.000}  &  \textit{0.003$\pm$0.000}  & \textit{0.054$\pm$0.002}  &  \textit{0.001$\pm$0.000} &  \textit{0.000$\pm$0.000} &  \textit{0.000$\pm$0.000} \\
\bottomrule
\end{tabular}}
\end{table*}

\begin{table*}[!ht]
\small 
\centering
\caption{MAP scores for the taxonomy enrichment methods for the Russian datasets. Numbers \textbf{in bold} show the best model within the category, \underline{\smash{\bf underlined}} numbers denote the best score across all the models.}
\label{tab:all_rus}
\resizebox{0.85\textwidth}{!}{
\begin{tabular}{l|cc|cc}
\toprule
\multicolumn{1}{c|}{\multirow{2}{*}{method}}                                     & \multicolumn{2}{c|}{nouns}                            & \multicolumn{2}{c}{verbs}         \\
\multicolumn{1}{c|}{}                                                            & \multicolumn{1}{c}{non-restricted} & restricted         & \multicolumn{1}{c}{non-restricted}  & restricted         \\
\midrule
\multicolumn{5}{c}{\bf Baseline}\\
\midrule
fastText \cite{bojanowski-etal-2017-enriching}                                                                 & \textbf{0.414$\pm$0.001}                            & \textbf{0.549$\pm$0.006}         & 0.296$\pm$0.004        & 0.389$\pm$0.011         \\
word2vec \cite{word2vec}                                                                & 0.263$\pm$0.001                            & 0.427$\pm$0.006         & \textbf{0.343$\pm$0.004}          & \textbf{0.445$\pm$0.013}         \\
\midrule
\multicolumn{5}{c}{\bf DWRank (Word Embeddings)}\\
\midrule
 fastText \cite{bojanowski-etal-2017-enriching}&\textbf{0.419$\pm$0.001}&\textbf{0.572$\pm$0.005}&\textbf{0.337$\pm$0.003}&\textbf{0.428$\pm$0.007}\\
 word2vec \cite{word2vec}&0.296$\pm$0.002&0.569$\pm$0.005&0.250$\pm $0.003&0.284$\pm$0.011\\
\midrule
\multicolumn{5}{c}{\bf DWRank (Meta-embeddings based on Word Embeddings)}\\
\midrule
concat (\textit{words})&0.422$\pm$0.001&0.589$\pm$0.005&0.351$\pm$0.004&0.426$\pm$0.009\\
SVD (\textit{words})&0.461$\pm$0.001&\textbf{0.600$\pm$0.005}&\textbf{\uline{0.426$\pm$0.005}}&\textbf{\uline{0.475$\pm$0.010}}\\
CAEME (\textit{words}) &0.400$\pm$0.001&0.561$\pm$0.005&0.342$\pm$0.003&0.416$\pm$0.008\\
AAEME (\textit{words}) &0.456$\pm$0.001&0.582$\pm$0.005&0.368$\pm$0.004&0.442$\pm$0.009\\
\midrule
CAEME triplet loss (\textit{words})&0.449$\pm$0.001&0.581$\pm$0.005&0.374$\pm$0.003&0.427$\pm$0.010\\
AAEME triplet loss (\textit{words})&\textbf{\uline{0.474$\pm$0.001}}&0.593$\pm$0.006&0.399$\pm$0.004&0.449$\pm$0.010\\
\midrule
\multicolumn{5}{c}{\bf DWRank (Graph embeddings)}\\
\midrule
GCN \cite{kipf2016semi}&0.183$\pm$0.001&0.306$\pm$0.005&0.220$\pm$0.003&0.287$\pm$0.009\\
GAT \cite{velickovic2018graph}&0.142$\pm$0.001&0.318$\pm$0.004&0.000$\pm$0.000&0.000$\pm$0.000\\
GraphSAGE \cite{hamilton2017inductive}&0.176$\pm$0.001&0.348$\pm$0.005&0.181$\pm$0.003&0.226$\pm$0.008\\
TADW \cite{yang2015network} & \textbf{0.417$\pm$0.001}&\textbf{0.562$\pm$0.005}&\textbf{0.328$\pm$0.003}&\textbf{0.423$\pm$0.008}\\
Poincaré \cite{nickel2017poincare} (top-5 fastText associates)&0.336$\pm$0.001&0.476$\pm$0.005&0.244$\pm$0.004&0.339$\pm$0.009\\
node2vec \cite{node2vec-kdd2016} (top-5 fastText associates)&0.343$\pm$0.002&0.477$\pm$0.005&0.226$\pm$0.003&0.322$\pm$0.010\\
HOPE \cite{ou2016asymmetric}& 0.000$\pm$0.000                              &  0.000$\pm$0.000        & 0.003$\pm$0.001        & 0.003$\pm$0.001         \\
\midrule
\multicolumn{5}{c}{\bf DWRank (Meta-embeddings based on Word and Graph Embeddings)}\\
\midrule
SVD (\textit{words} + node2vec)&0.367$\pm$0.001&0.521$\pm$0.005&0.252$\pm$0.003&0.351$\pm$0.010\\
CAEME (\textit{words} + node2vec)&0.370$\pm$0.001&0.533$\pm$0.005&0.267$\pm$0.003&0.362$\pm$0.010\\
AAEME (\textit{words} + node2vec)&0.373$\pm$0.001&0.529$\pm$0.005&0.272$\pm$0.003&0.358$\pm$0.010\\
\midrule
SVD (\textit{words} + TADW)&\textbf{0.469$\pm$0.001}&\textbf{\uline{0.604$\pm$0.006}}&\textbf{0.394$\pm$0.005}&\textbf{0.455$\pm$0.010}\\
CAEME (\textit{words} + TADW)&0.429$\pm$0.001&0.571$\pm$0.005&0.349$\pm$0.003&0.437$\pm$0.009\\
AAEME (\textit{words} + TADW)&0.461$\pm$0.001&0.584$\pm$0.005&0.362$\pm$0.004&0.439$\pm$0.009\\
\midrule
SVD (\textit{words} + GCN)&0.395$\pm$0.001&0.554$\pm$0.005&0.291$\pm$0.004&0.356$\pm$0.009\\
CAEME (\textit{words} + GCN)&0.389$\pm$0.001&0.544$\pm$0.005&0.302$\pm$0.003&0.381$\pm$0.008\\
AAEME (\textit{words} + GCN)&0.386$\pm$0.001&0.545$\pm$0.006&0.295$\pm$0.004&0.365$\pm$0.008\\
\midrule
SVD (\textit{words} + GraphSAGE)&0.410$\pm$0.001&0.603$\pm$0.005&0.336$\pm$0.004&0.426$\pm$0.009\\
CAEME (\textit{words} + GraphSAGE)&0.321$\pm$0.001&0.541$\pm$0.005&0.266$\pm$0.004&0.345$\pm$0.007\\
AAEME (\textit{words} + GraphSAGE)&0.409$\pm$0.001&0.577$\pm$0.006&0.323$\pm$0.004&0.419$\pm$0.009\\
\midrule
\multicolumn{5}{c}{\bf State-of-the-art Approaches}\\
\midrule
WBSR (Top-1 RUSSE'2020 for nouns)            & \multicolumn{1}{c}{\textbf{0.393$\pm$0.002}}              & \textbf{0.552$\pm$0.005}          & 0.293$\pm$0.004               & 0.428$\pm$0.010\\
WBSR, no search engine features                                                               & \multicolumn{1}{c}{0.369$\pm$0.002}              & 0.497$\pm$0.005          & 0.267$\pm$0.004               & 0.387$\pm$0.009\\
Top-1 RUSSE'2020 for verbs: \cite{dale2020russe}                                                                 & \multicolumn{1}{c}{0.288$\pm$0.001}              & 0.418$\pm$0.006         & \textbf{0.341$\pm$0.004}              & \textbf{0.452$\pm$0.012}\\\midrule

hypo2path \cite{cho-etal-2020-leveraging}         &        0.061$\pm$0.000                     &   0.097$\pm$0.002       &       0.137$\pm$0.003   &      0.174$\pm$0.009     \\ 
hypo2path rev \cite{cho-etal-2020-leveraging}              &      0.246$\pm$0.001                     &    0.342$\pm$0.006     &     0.151$\pm$0.003  &    0.194$\pm$0.008      \\
hypo2path rev transformer \cite{cho-etal-2020-leveraging}        &         0.234$\pm$0.001                     &    0.331$\pm$0.004       &     0.152$\pm$0.003     &    0.201$\pm$0.008      \\ 
\midrule
\textit{TaxoExpan} \cite{shen2020taxoexpan}      &  \textit{0.007$\pm$0.000}  &  \textit{0.006$\pm$0.001}  &  \textit{0.009$\pm$0.001} &  \textit{0.008$\pm$0.002} \\
\bottomrule
\end{tabular}
}
\end{table*}

\end{document}